\documentclass[article,3p,authoryear,times]{elsarticle}

\usepackage{graphicx}
\usepackage{subcaption}
\usepackage[fleqn]{amsmath}
\usepackage[]{amsfonts}
\usepackage[]{amsthm}
\usepackage[]{amssymb}
\usepackage{empheq}
\usepackage{rotating,tabularx}
\usepackage{multirow}
\usepackage{geometry}
\usepackage[export]{adjustbox}
\usepackage{wrapfig}
\usepackage{siunitx}
\usepackage{xcolor}
\usepackage{multicol}
\usepackage{siunitx}

\usepackage{caption}
\usepackage{graphicx, stmaryrd, url}

\usepackage{bm}
\usepackage{glossaries}
\usepackage{amsmath}
\usepackage{float}

\def\W{\mathbf{W}}

\def\I{\mathbf{I}}

\def\F{\mathbf{F}}
\def\P{\mathbf{P}}
\def\X{\mathbf{X}}
\def\Q{\mathbf{Q}}
\def\K{\mathbf{K}}
\def\V{\mathbf{V}}
\def\B{\mathbf{B}}

\def\G{\mathbf{G}}
\def\E{\mathbf{E}}
\def\D{\mathbf{D}}

\newglossaryentry{lengthtree}%
{%
  name={$L_t$},
  text={L_t},
  description={description here}
  sort={L}
}

\usepackage{ifthen}
\newboolean{finalversion}
\setboolean{finalversion}{true}
\setboolean{finalversion}{false}

\usepackage{soulutf8}
\usepackage{color}
\usepackage{ragged2e}
\usepackage{xargs}
\newcounter{commentnumber}
\newcommandx{\comment}[3][1=,2=]{%
  \hl{$^{\arabic{commentnumber}}$}\stepcounter{commentnumber}%
  \hl{#1}%
  \ifthenelse{\equal{#2}{}}%
  {}%
  {\hl{$\rightarrow$#2}}%
  \ifthenelse{\equal{#1}{}\AND\equal{#2}{}}%
  {}%
  {\hl{: }}
  \hl{#3}}
\iffinalversion
\renewcommandx{\comment}[3][1=,2=]{}
\fi

\usepackage[mathlines]{lineno}

\usepackage[english]{babel}
\usepackage[T1]{fontenc}

\usepackage{hyperref} 
\usepackage{graphicx}
\usepackage{tikz}
\usepackage{pstricks}
\usepackage{xstring}
\usepackage{multirow}
\usepackage{booktabs,tabularx}

\definecolor{darkblue}{rgb}{0,0,0.5}

\hypersetup{colorlinks=true,linkcolor=black,citecolor=darkblue,urlcolor=darkblue} 

\newcommand{\newref}[2]{\hyperref[#2]{#1~\ref*{#2}}} 

\begin{document}

\begin{frontmatter}

\title{From Optimization to Prediction: Transformer-Based Path-Flow Estimation to the Traffic Assignment Problem}

\author[GRETTIA,BERKELEY]{Mostafa Ameli\corref{cor}}
\author[GRETTIA]{Sulthana Shams}
\author[GRETTIA]{Van Anh Le}
\author[BERKELEY]{Alexander Skabardonis}

\address[GRETTIA]{Univ. Gustave Eiffel, COSYS, GRETTIA, Paris, France}
\address[BERKELEY]{Institute of Transportation Studies, University of California, Berkeley, USA}

\begin{abstract}

The traffic assignment problem is essential for traffic flow analysis, traditionally solved using mathematical programs under the Equilibrium principle. These methods become computationally prohibitive for large-scale networks due to non-linear growth in complexity with the number of OD pairs. This study introduces a novel data-driven approach using deep neural networks, specifically leveraging the Transformer architecture, to predict equilibrium path flows directly. By focusing on path-level traffic distribution, the proposed model captures intricate correlations between OD pairs, offering a more detailed and flexible analysis compared to traditional link-level approaches. The Transformer-based model drastically reduces computation time, while adapting to changes in demand and network structure without the need for recalculation. {Numerical experiments conducted on the Manhattan-like synthetic network, the Sioux Falls network, the Eastern-Massachusetts network, and the Anaheim network demonstrate that the proposed model is orders of magnitude faster than conventional optimization. It efficiently estimates path-level traffic flows in multi-class networks, reducing computational costs and improving prediction accuracy by capturing detailed trip and flow information. The model also adapts flexibly to varying demand and network topology conditions, supporting traffic management and enabling rapid `what-if' analyses for enhanced transportation planning and policy-making.}

\end{abstract}


\end{frontmatter}



\section{Introduction}
The \textit{Traffic Assignment Problem (TAP)} is a process of determining the propagation of flows over the transportation network. 
The goal is to calculate the network state, given the travel demand between various \textit{origin-destination (OD)} pairs and the network's capacity constraints \citep{sheffi}. 
Traditionally, this problem is solved through mathematical programs under the \textit{User Equilibrium (UE)} principle, which assumes drivers possess perfect information and make fully rational choices \citep{doi:10.1680/ipeds.1952.11259}.
Despite potential deviations from reality, this approach consistently provides reasonable solutions to the traffic assignment problem \citep{bar-gera02,JAFARI17}. However, the computation for determining optimal solutions in large traffic networks is prohibitively costly. 
This is because the problem's complexity grows non-linearly with the increase in the number of OD pairs and directly depends on feasible paths.
When the size of the network (the number of links and nodes in a representative graph) increases, allowing us to explore more paths, the number of feasible paths also increases, and the OD demand matrix may grow accordingly, leading to a non-linear increase in computation time \citep{Patriksson}. 
Because changing the input (demand and graph) results in a new problem to be solved, it requires re-calculation. Therefore, there is a critical need to find another way to capture the complex nature of traffic flow and demand in large networks in order to use the already-solved problem to estimate the solution to the new problem. 

Mathematically, the UE, is typically calculated using optimization methods such as the gradient method or fixed-point algorithm \citep{Heqing20, Mehrzad21, LIU2023104085}, e.g., {Frank-Wolfe algorithm \citep{fukushima1984modified}}. {While these optimization-based approaches provide near-optimal solutions, they require solving the entire network problem from scratch for each `what-if' scenario, making them computationally expensive and impractical for real-time or large-scale applications.} However, with the presence of data-driven approaches and the application of machine learning, an alternative solution to this problem could be prediction instead of optimization.
In other words, we can employ supervised learning to learn the optimal solution and predict it given a new demand or a new graph. 
Recently, a few studies have tried to estimate the link flow solution of UE using machine learning \citep{fan23, rahman2022datadriven, liu24, liu2025scalable, LIU2025128072, hu2025use} with graph neural networks (GNNs) emerging as a particularly popular approach for utilising network topology to capture spatial dependencies. {For example, \cite{rahman2022datadriven} and \cite{liu2025scalable} use graph convolutional neural networks (GCN) to forecast traffic flow patterns}, while \cite{liu24, hu2025use}  employed graph attention networks (GATs) to accelerate assignment across large-scale planning scenarios.
While promising, GNNs are constrained by their reliance on local message passing, which limits their ability to model long-range dependencies. Deeper stacking can address this, but at the cost of higher computation and over-smoothing of features.

Other approaches have pursued surrogate formulations of the traffic assignment problem. For instance, \cite{zhang2025low} introduced a low-rank approximation for path-based network models, generating hypothetical path sets based on flows rather than topological structure. {\cite{LIU2023104085} propose an end-to-end learning framework that embeds the traffic equilibrium problem as a Variational Inequality (VI) layer, producing flows through an iterative equilibrium procedure.} Yet these methods remain tied to surrogate calibration or traditional iterative equilibrium procedures. 
While Deep Neural Networks (DNNs) have proven effective in domains such as computer vision and NLP \cite{wang22}, and are increasingly applied to transportation forecasting, the use of sequence-based architectures, remains largely unexplored in TAP \citep{WEN2023119587, XUE2025110263, JIA2025104979}. {Recent studies have applied attention mechanisms to enhance sequential modeling in traffic forecasting, such as hierarchical attention-based LSTM frameworks for network-level traffic state prediction \cite{Zhang_2024}, demonstrating the ability of attention to capture multi-scale temporal dependencies. Building on this, Transformer-based architectures have been explored in related transportation applications involving spatiotemporal time series prediction, including trajectory and motion forecasting \cite{ref_LGT} and traffic flow prediction \cite{CNNTransformer}, demonstrating their effectiveness in modeling long-range dependencies and global patterns in traffic data. Despite these advancements, the application of Transformers to the TAP remains largely unexplored. }

To address these limitations, we introduce a Transformer-based framework for TAP that employs global self-attention, enabling each element of the input sequence to dynamically weigh its relevance to all others and capture complex, long-range dependencies across the network. The attention mechanism allows the model to selectively prioritize OD pairs with nonzero demand or links with critical features while maintaining global context, enhancing learning efficiency. In contrast, prior TAP models using GAT-based encoder-decoder architectures restrict attention to local neighborhoods and decode flows through MLP layers, limiting their ability to model global interactions \citep{hu2025use}. The proposed model offers both computational efficiency and robustness, providing a scalable alternative to traditional iterative equilibrium solvers. To the best of our knowledge, this represents the first TAP framework to employ a Transformer-based sequence learning architecture integrating both encoder and decoder global attention.


{Travelers make decisions at the path level, and equilibrium arises from their collective route-choice interactions. Most existing approaches emphasize link-level predictions, leaving path-level demand propagation underexplored, despite its importance for capturing traveler behavior. In contrast, the proposed path-based Transformer learns the equilibrium directly in path-flow space, aligning with the game-theoretic interpretation of TAP. In doing so, the model  reformulates TAP as a traveller decision-making process rather than only a traffic prediction problem.  By providing detailed path-level flows, it captures the interactions among OD pairs that give rise to equilibrium and provides an interpretable mapping between demand and network conditions. This reformulation addresses two persistent challenges in current practice, the computational cost of iterative solvers and the lack of travelers choice representation in link-based models. Trained on feasible equilibrium data, the model inherently preserves flow conservation and capacity constraints while supporting scenario-based evaluations such as network or demand perturbations, enabling generalization across unseen conditions. This formulation also establishes a foundation for route-choice guidance or decision-support applications in TAP.}

To summerize, the major contributions of this work are as follows: (i) A Transformer-based model is proposed for directly learning and predicting UE path flow distributions in both single- and multi-class traffic networks; (ii) The model incorporates both network topology and detailed OD demand information, allowing it to generalize to unseen demand patterns and network changes; (iii) Our approach eliminates the computationally intensive optimization step, providing solutions orders of magnitude faster than traditional methods; and (iv) The model supports robust '`what-if" scenario analysis under extreme perturbations, where link-removal and link addition based topology changes, missing-link observation cases, as well as multi-class and large-scale network applications particularly in terms of OD demand are considered jointly with incomplete OD demand.

The paper is structured as follows: \newref{Section}{sec:literature} presents a summary of relevant literature in the field of traffic assignment and deep learning to better position this paper. \newref{Section}{sec_compare_nn} compares the proposed transformer-based model with other state-of-the-art TAP models. \newref{Section}{sec:architecture} formalizes the problem and describe detail model architecture. Numerical experiments are conducted on the Manhattan-like network, Sioux Falls, and Eastern-Massachusetts network in \newref{Section}{sec:numerical}. Finally, \newref{Section}{sec:conclude} concludes the paper.

\section{Related literature}\label{sec:literature}

The problem of traffic flow prediction and traffic assignment has been extensively studied, with solutions that could be categorized into analytical and data-driven approaches.
Analytical methods rely on mathematical formulations and traffic theories to model and solve traffic problems. 

{Beginning with UE principle of \cite{doi:10.1680/ipeds.1952.11259} and the convex optimisation formulation of \cite{beckman1956cb}, these approaches laid the foundation of the TAP.} To reduce the need for full path enumeration, some analytical approaches developed link-based algorithms, which directly solve for link flows \citep{fukushima1984modified}, as well as column and bush-based methods, which aim to find only promising subsets of paths to improve efficiency \citep{column_based_TAP, bar-gera02, bush_based_TAP}. Despite these advances, TAP remains computationally intensive for large networks, where evaluating thousands of `what-if' scenarios can take days. This motivates the shift toward supervised machine learning models trained on solver-generated solutions, offering significantly faster inference while retaining the accuracy of the underlying optimization framework.

These approaches integrate machine learning with optimization, where a solver guides training or inference.   Models are trained on solver-generated optimal flows to predict link or path flows efficiently \citep{su_dl_optimal_training,LIU2023104085,rahman2022datadriven}.
These data-driven approaches employed the power of machine learning to extract patterns from vast amounts of traffic data \citep{WEN2023119587, kipf2017}. 
To predict the optimal solution,  machine learning approaches learn path flow distribution under UE conditions.
Various machine learning models have been developed for traffic flow prediction, including Graph Neural Networks (GNNs), Heterogeneous Graph Neural Networks (HGNNs), and other deep learning frameworks like Transformer and Recurrent Neural Network (RNN). 
GNNs and HGNNs excel at representing spatial dependencies and interactions within the network structure \citep{PhysRevResearch}, while Transformers offer flexibility and efficiency in handling large-scale data with complex spatial-temporal relationships \citep{WEN2023119587}. 
As we explore these approaches, it becomes evident that each model offers unique advantages and limitations, making the choice of model highly dependent on the specific characteristics of the traffic prediction tasks.
The following sections review studies related to GNN-based methods and Transformer models as deep learning approaches. We examine their advantages and drawbacks to determine which method is more reliable for our approach.

\subsection{Graph Neural Networks (GNNs)} \label{subsec2.1}
GNN is a powerful framework for learning representations of graphs \citep{kipf2017}. 
They operate through a neighborhood aggregation scheme, where each node's representation is computed by sampling and aggregating features from its neighboring nodes. This process allows GNNs to capture the structural and relational information inherent in graph data \citep{kipf2017, NIPS2017}. 
\cite{Nishi2018} developed a traffic signal control algorithm based on graph convolutional networks (GCN) and reinforcement learning. 
\cite{yu2018urban} introduced a spatio-temporal GCN to predict urban traffic flows, capturing both spatial and temporal correlations. 
However, the grid-based convolutional neural network (CNN) models cannot fully capture the intricate structures of transportation systems. 
Besides, these models inadequately account for variations in the transportation network, such as reductions in link capacity due to traffic accidents or lane closures for maintenance.

Recently \cite{liu24, LIU2025128072} proposed a framework utilizing HGNN for end-to-end traffic assignment and traffic flow learning.
They integrated spatial information into the graph representation of transportation networks, enabling feature extraction and aggregation for traffic flow predictions.
\cite{jiang2021graph} developed a traffic flow prediction model using graph attention networks to capture spatial dependencies in traffic networks. 
\cite{Guo2019} used an attention mechanism to capture the dynamics of spatial dependencies and temporal correlations.
A recent work, \cite{hu2025use}, {introduced an encoder–decoder transformer-inspired GAT framework} that leverages attention mechanisms to map network features to link or path flows. However, their framework still relies on surrogate model calibration by equilibrium-based solution algorithms. We propose to directly solve for path flow even in multi-class network. While these models have demonstrated promising performance, they still face scalability challenges due to computational complexity and memory requirements, as capturing long-range dependencies requires stacking multiple layers \citep{Li_GNN, over_smoothing, zhou2020graph}.

\subsection{Deep Learning (DL) models} \label{subsec2.2}

\cite{LIU2023104085} introduced an end-to-end learning framework for user equilibrium (UE) using implicit neural networks. Their approach directly learns travel choice preferences from data and captures equilibrium conditions with a neural network-based variational inequality. However, their method does not approach equilibrium on the flow side to determine the network state. Secondly, they still rely on iterative optimization to calculate solutions rather than directly estimating them, resulting in scalability issues.

Besides the calculation of UE by optimization approaches, there are multiple deep learning approaches for traffic estimation, which can be related to estimating the state of the network. 
RNNs have been widely used to model temporal dependencies in traffic flow data for prediction. Studies like those by \cite{bruno2020lstm} and \cite{wu2023lstm} have shown the ability of Long Short Term Memory (LSTM) networks, a type of RNN, to capture long-term and short-term dependencies in traffic patterns. {More recently, hierarchical attention-based extensions of LSTM models have been proposed for network-level traffic state prediction \cite{Zhang_2024}, effectively capturing multi-scale temporal dependencies and improving the forecasting of congestion patterns.}
However RNNs are recursive and cannot parallelize all sequences in parallel, and they are weak for variable-length data \citep{MA2015187, BOGAERTS202062}.
\cite{fan23} had built a CNN-based model to estimate link flow from incomplete OD data. 
However, this model was applied on a small-sized dataset of 1000 data points and it does not address the estimation tasks of path flow distribution.

Transformer architectures have shown promise by providing insights into model decisions through attention mechanisms, enabling stakeholders to understand factors influencing traffic patterns \citep{mandalis2025transformer}. They have been increasingly applied to traffic flow forecasting tasks \citep{WEN2023119587, XUE2025110263, JIA2025104979}. For instance, \citet{WEN2023119587} proposed a hybrid model that integrates convolutional neural networks (CNNs) with a Transformer framework to capture both local spatial features and long-range dependencies in traffic flow prediction. However, this approach has some limitations: The proposed architecture is complex, making it infeasible for larger scales because the computation time increases non-linearly. Additionally, the model may struggle to generalize well to different supplies and demands not seen during training, potentially limiting its applicability in real-world traffic management. 


Thus, while Transformer models have been widely applied in the context of short-term traffic forecasting, their application to solving TAP remains largely unexplored. This study contributes to this gap by investigating the use of Transformer-based architectures to directly learn path flow distributions under UE conditions.

%

\section{Comparison Between the Proposed  Model and Other Neural Network Models}\label{sec_compare_nn}

\begin{table}[!h]
\centering
\caption{Comparison of ML-based approaches in the literature. \textbf{Column Descriptions:} \textbf{Baseline OD Data} indicates the availability of OD demand; in our work, all perturbations are applied on top of a 30\% missing OD training data, reflecting realistic partial demand information. \textbf{Perturbations} include: \textbf{Multi-class} (predicting flows for multiple vehicle classes); \textbf{Link Removal} (simulating abrupt events such as road closures; links removed during testing are not part of training, testing model adaptability; in some studies, partial capacity variations are used instead of full removal); \textbf{Link Missing} (links absent from network observations, simulating incomplete network knowledge); and \textbf{OD Missing} (partial OD demand observed). \textbf{Conservation Handling} is split into two columns: \textbf{Link Flow} indicates link-level or node-level conservation, either implicit via path aggregation or diffusion flow modeling across adjacent nodes or explicitly enforced through a loss term; \textbf{OD} indicates OD-demand conservation, whether it is implicit by design, enforced via post-processing, or not considered. }
\resizebox{\textwidth}{!}{%
\begin{tabular}{lccccccccc}
\hline
\multirow{2}{*}{\begin{tabular}[c]{@{}l@{}}  \textbf{Model Type}\\  \textbf{({\color{blue}Reference})}\end{tabular}}                 & \multicolumn{2}{c}{\textbf{Model Output}}                                                                                                             & \multirow{2}{*}{\textbf{\begin{tabular}[c]{@{}c@{}}Baseline\\ OD\\ Data\end{tabular}}} & \multicolumn{4}{c}{\textbf{Perturbations}}                                                                                                                                                                                                                                                                   & \multicolumn{2}{c}{\textbf{Conservation Handling}}                                                                                               \\ \cline{2-3} \cline{5-8} 
                                                                                                   & \multicolumn{1}{c}{\textbf{\begin{tabular}[c]{@{}c@{}}Link\\ Flow\end{tabular}}} & \textbf{\begin{tabular}[c]{@{}c@{}}Path\\ Flow\end{tabular}} &                                                                                        & \multicolumn{1}{c}{\textbf{\begin{tabular}[c]{@{}c@{}}Multi-\\ class\end{tabular}}} & \multicolumn{1}{c}{\textbf{\begin{tabular}[c]{@{}c@{}}Link\\ Removal\end{tabular}}} & \multicolumn{1}{c}{\textbf{\begin{tabular}[c]{@{}c@{}}Link\\ Missing\end{tabular}}} & \textbf{\begin{tabular}[c]{@{}c@{}}OD\\ Missing\end{tabular}} & \multicolumn{1}{c}{\textbf{\begin{tabular}[c]{@{}c@{}}Link\\ Cons.\end{tabular}}} & \textbf{\begin{tabular}[c]{@{}c@{}}OD\\ demand\end{tabular}} \\ \hline
\begin{tabular}[c]{@{}l@{}}Implicit-layer NN\\ \citep{LIU2023104085}\end{tabular} & \multicolumn{1}{c}{\checkmark}                                    &                                                              & Fixed given 100 \%                                                                     & \multicolumn{1}{c}{--}                                                              & \multicolumn{1}{c}{--}                                                              & \multicolumn{1}{c}{\checkmark}                                       & \checkmark                                     & \multicolumn{1}{c}{Implicit}                                                      & --                                                 \\ 
\begin{tabular}[c]{@{}l@{}}GCN\\ \citep{rahman2022datadriven}\end{tabular}        & \multicolumn{1}{c}{\checkmark}                                    &                                                              & Fixed given 100 \%                                                                     & \multicolumn{1}{c}{--}                                                              & \multicolumn{1}{c}{Capacity Variation}                    & \multicolumn{1}{c}{--}                                                              & --                                                            & \multicolumn{1}{c}{Implicit}                                                      & --                                                           \\ 
\begin{tabular}[c]{@{}l@{}}HetGAT\\  \citep{liu24}\end{tabular}                   & \multicolumn{1}{c}{\checkmark}                                    &                                                              & Fixed given 100 \%                                                                     & \multicolumn{1}{c}{--}                                                              & \multicolumn{1}{c}{Capacity Variation}                    & \multicolumn{1}{c}{--}                                                              & \checkmark                                     & \multicolumn{1}{c}{Explicit}                                                      & --                                                           \\ 
\begin{tabular}[c]{@{}l@{}}GAT-L, GAT-P\\ \citep{hu2025use}\end{tabular}          & \multicolumn{1}{c}{\checkmark}                                    & \checkmark                                    & Fixed given 100 \%                                                                     & \multicolumn{1}{c}{--}                                                              & \multicolumn{1}{c}{--}                                                              & \multicolumn{1}{c}{--}                                                              & \checkmark                                     & \multicolumn{1}{c}{Implicit}                                                      & Post-process                                                           \\ 
\begin{tabular}[c]{@{}l@{}}M-HetGAT\\  \citep{LIU2025128072}\end{tabular}         & \multicolumn{1}{c}{\checkmark}                                    &                                                              & Fixed given 100 \%                                                                     & \multicolumn{1}{c}{\checkmark}                                       & \multicolumn{1}{c}{\checkmark}                                       & \multicolumn{1}{c}{--}                                                              & \checkmark                                     & \multicolumn{1}{c}{Explicit}                                                      & --                                                           \\ 
\begin{tabular}[c]{@{}l@{}}Transformer-based\\ ({\color{blue}This study})\end{tabular}                             & \multicolumn{1}{c}{\checkmark}                                    & \checkmark                                    & Missing (30\%)                                                                         & \multicolumn{1}{c}{\checkmark}                                       & \multicolumn{1}{c}{\checkmark}                                       & \multicolumn{1}{c}{\checkmark}                                       & \checkmark                                     & \multicolumn{1}{c}{Implicit}                                                      & Implicit                                                     \\ \hline
\end{tabular}}\label{compare_nn}
\end{table}

As discussed in previous sections, the application of neural networks to TAP remains relatively underexplored. Table \ref{compare_nn} compares our approach with other state-of-the-art neural network based traffic assignment models in terms of model outputs, handling of perturbations, and flow conservation strategies. Beyond introducing a new architecture, this study empirically demonstrates that a Transformer-based sequence learning formulation can reliably replicate User Equilibrium conditions, preserve OD-level conservation, and sustain performance under simultaneous demand and topology perturbations. 

In TAP, flow conservation is essential for physically realistic and equilibrium-consistent predictions, ensuring that OD demand is met and that traffic entering and leaving each node balances correctly. Path-flow formulations, including the approach proposed in this study,  respect link flow conservation through path-to-link aggregation, providing feasible equilibrium-consistent solutions \citep{LIU2023104085, hu2025use}. In contrast, link-based approaches often require explicit enforcement, such as adding node-based conservation terms to the loss function \cite{LIU2025128072}, while GCN-based model by \cite{rahman2022datadriven} approximates conservation by modelling flow diffusion across neighbours. 

While link or path-level flow conservation ensures locally feasible flows, it does not necessarily guarantee that OD demand is satisfied. To address this, \cite{hu2025use} applied a post-processing step to adjust path flows and enforce OD demand consistency. In contrast, our path-based Transformer model, trained on ground-truth path flows generated under equilibrium conditions, predicts flows that inherently sum to each OD pair’s demand, ensuring OD-level conservation without any additional adjustment. Because OD conservation is implicit and enforced through formulation in Equation \ref{eq:3}, no iterative enforcement or adjustment is needed at inference, making predictions potentially faster.

Our evaluation deliberately incorporates extreme network perturbations, such as link removal and missing link observation data, as they provide a rigorous test of the model’s generalization capacity and robustness under realistic and challenging conditions. Our model demonstrates robust performance under varying levels of OD data missing levels and link-removal-based topology changes, even without retraining.
Further, the studies examine the robustness of the model by either supply or demand side disruptions in isolation, and do not account for simultaneous demand and supply changes, demonstrating more practical and realistic testing conditions. This study addresses these gaps by performing `what-if' analyses under scenarios involving joint supply-demand perturbations. Missing OD demand is adopted as a baseline perturbation in this work, as it reflects a common and practically unavoidable source of uncertainty in real-world applications. Building on this baseline, we introduce additional perturbations such as link removal or missing link data. 
Lastly, all of the aforementioned studies focus on single-class network graphs, with the exception of \citet{LIU2025128072}, which extends the model to multi-class traffic scenarios. While their methodological framework differs from ours, their choice of test cases is closely aligned with our work. We further extend the evaluation by jointly considering network perturbations with incomplete OD data, demonstrating enhanced adaptability. 

This work introduces a Transformer-based sequence learning framework for TAP that directly estimates path-level equilibrium flows in single and multi-class networks. The model inherently satisfies link and OD-level flow conservation, handles joint supply-demand perturbations, and adapts to changes in network topology and demand patterns. By providing a computationally efficient surrogate for traditional optimization solvers, it supports realistic `what-if' analyses and demonstrates the practical potential of sequence-based architectures for large-scale traffic assignment.

\section{Transformer-based model for optimal path flow prediction} \label{sec:architecture}
\subsection{Problem formulation}
As mentioned before, the goal is to estimate the optimal solution of path flow distribution from given OD demands and the graph instead of solving it. 
In mathematical terms, we consider a network represented as a graph $G = (V, E)$, where $V$ stands for the set of nodes with size $N$, and $E$ for the set of links. 
Let $R$ represent the set of OD pairs. 
For each OD pair, denoted by $r\in R$, where $r = (v_1, v_2), \forall v_1, v_2 \in V$, vehicles of class $z \in Z$ can travel via several finite and nonempty feasible paths of path set $\P_r$. Let $n$ denote the number of classes in the network.
Each set $\P_r\in \P^k$ encompasses $k$ feasible paths for each OD pair $r$. 
Furthermore, let $\X$ represents OD demand matrix, $x^z_r \in \X$ represents the demand of OD pair $r$ traveled by class $z$, and $f^{r,z}_{\P_r} \in \F$ represents paths flow distributions of OD pair $r$ on path set $\P_r$, with class $z$. 
Each set $f^{r,z}_{\P_r}$ encompasses $k$ flow values for every OD pair $r$.
 
Following the UE principle, the shortest path is the path that has the minimum cost.
In the UE solution, all used paths must be at the minimum possible cost (or equal the shortest path). Mathematically, it is equivalent to the following conditions:
\begin{equation}\label{eq:1}
f^{r, z}_p(c^{r, z}_p - u^z_r) = 0 \quad \forall p \in \P_r, r \in R, z \in Z
\end{equation}
\begin{equation}\label{eq:2}
c^{r, z}_p - u^z_r \ge 0  \quad \forall p \in \P_r, r \in R, z \in Z
\end{equation}
\begin{equation}\label{eq:3}
\sum_{p \in \P_r} f^{r, z}_p = x^z_r \quad \forall r \in R,\ \forall z \in Z
\end{equation}

\begin{equation}\label{eq:4}
f^{r, z}_p \ge 0, c^{r, z}_p \ge 0, u^z_r \ge 0 \quad \forall p \in \P_r, r \in R, z \in Z
\end{equation}
\noindent where $c^{r, z}_p$ denotes the cost of path $p$ with class $z$ for OD pair $r$, which equals the sum of the cost of all links in that path (\newref{Equation}{eq:pathcost}), and $u^z_r$ denotes the minimum path travel cost of OD pair $r$ with class $z$ (\newref{Equation}{eq:mincost}):
\begin{equation}\label{eq:pathcost}
c^{r,z}_p \;=\; \sum_{e \in E} \Delta^{(r)}_{p,e}\, c^{z}_e,
\qquad
\Delta^{(r)}_{p,e} \;=\; \mathbf{1}\{ e \in p \}.
\end{equation}
\begin{equation}\label{eq:mincost}
    u^z_r = \min_{\forall p \in \P_r} c^{r, z}_p \quad \forall r \in R, z \in Z
\end{equation}\vspace{0.2cm}
\noindent where $\Delta^{(r)}_{p,e}\!\in\!\{0,1\}$ indicates whether link $e$ lies on path $p$ for OD pair $r$, and $c^{z}_e$ is the class-$z$ link travel time computed via a BPR function:
\begin{equation}\label{eq:bpr}
c^{z}_e \;=\; t^{z}_{e,0}\!\left(1+0.15\left(\frac{\sum_{z'\in Z} v^{z'}_e}{C_e}\right)^{4}\right),
\end{equation}
where $t^{z}_{e,0}$ is the free-flow travel time, $C_e$ is the link capacity and $v^{z'}_e=\sum_{r\in R}\sum_{p\in \P_r}\Delta^{(r)}_{p,e}\,f^{r,z'}_p$ denotes the class-specific link flow aggregated from path flows.

The ultimate goal is to find $\F^*$, the optimal path flow distribution for UE condition defined by \cite{doi:10.1680/ipeds.1952.11259}. 
Each set of path flow distribution ${f^{r, z}_{\P^r}}^* \in \F^*, \forall r,  \forall z$ is the optimal solution for OD pair $r$ on path set $p_r$ with class $z$.  
\newref{Equation}{eq:1} shows that for all classes $z \in Z$, the paths connecting any OD pair $r$ can be categorized into two groups: those that carry flow, where the travel time $c^{r, z}_p$ matches the minimum OD travel time $u^z_r$, and those that do not carry flow, where the travel time is greater than or equal to the minimum OD travel time.
\newref{Equation}{eq:2} ensures that $u^z_r$ is the minumum path cost.
\newref{Equation}{eq:3} ensures that all the demand is served.
\newref{Equation}{eq:4} ensures the positivity of variables.
However, this UE solution is only applicable when all drivers have perfect knowledge of the network state. In reality, it is not the case, so normally there will be a delay in the network. In the literature, this imperfection of knowledge is represented by stochastic terms that lead to other equilibrium terms such as Stochastic User Equilibrium (SUE) or Boundedly Rational UE (BRUE) \citep{Ameli2020}.

The model proposed in this study aims to learn a function $\mathcal{F}(.)$ that maps $N^2 \times n$ instances of OD demand matrix $\X$ and $N^2 \times k$ instances of feasible path set $\P$ in graph $G$, to $N^2\times kn$ instances of optimal path flow distribution $\F^*$,
defined as \newref{Equation}{eq:prob_form}:
\begin{equation}\label{eq:prob_form}
\mathcal{F}(G, \X, \P) = \widehat{\F}^*_{G, \X},
\end{equation}
where $\widehat{\F}^*_{G, \X}$ is the predicted path flow distribution of the optimal solution $\F^*$ for graph $G$ with demand $\X$.

\subsection{Model architecture}

To achieve our goal, we developed a transformer-based framework that includes both Encoder and Decoder blocks.
\newref{Figure}{fig:model-flow} presents the overview of our framework, which includes four main parts: data preprocessing, Encoder, Decoder, and prediction. 

\begin{figure}[!b]
    \centering
    \includegraphics[width=0.8\textwidth]{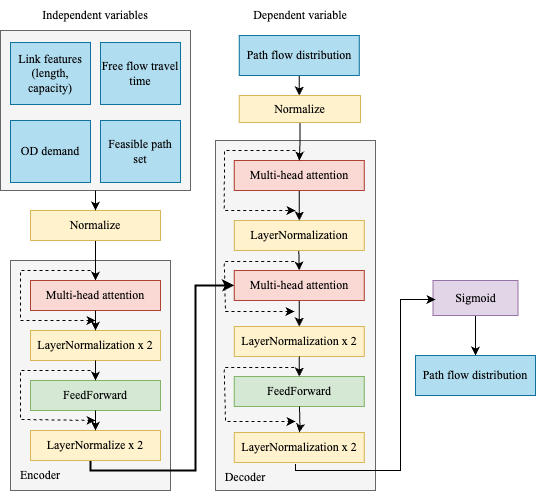}
    \caption{Model architecture }
    \label{fig:model-flow}
\end{figure}

Since the data is graph-based, where the graph and OD matrix are conditional inputs, we aim to estimate the optimal solution when we have variations in the graph and demand profile. An Encoder-Decoder architecture offers a robust framework for capturing complex dependencies \citep{transfer_learning}. The Encoder processes the input sequence (a series of OD pairs) and encodes this information into a context vector. The Decoder then uses this context vector to generate the output sequence (path flow distribution).
The Encoder-Decoder model can handle varying lengths of input and output sequences \citep{ghaderi2017deep}, which is particularly useful in transportation networks where the number of nodes and links can vary. This flexibility is advantageous for the `what-if' analysis.

In Transformer architectures, the attention mechanism is utilized in both the Encoder and Decoder blocks to enable the model to selectively focus on the meaningful values, and ignore the less meaningful values \citep{Attention}. 
The selective focusing in this problem means the model only focuses on the OD pairs that have a demand higher than zero ($x^z_r > 0$) or where there are features of the link between that pair, thereby enhancing the model's overall performance.
Below, we describe the architecture in detail, highlighting how the data is processed through each stage of the model.

\subsubsection{Data Preprocessing} 

To learn the graph, we need to ensure that all features are on the same scale. 
Since the raw data is not normalized, we perform a preprocessing step to encode the raw features into a lower-dimensional representation. 
The input tensor is constructed from four tensors. The first tensor stores graph information (link features), including the length and capacity of each link. The second tensor stores free-flow travel time for each link and each class. The third tensor contains the OD demand matrix for each class. The fourth tensor holds the encoded feasible paths information for each OD pair.

Let $m$ denotes the number of link features. For $n$ number of classes and $k$ feasible paths in the network, the dimension $a$ of the input tensor is calculated as: $a = m + 2n + k$.
All the notations in this section are presented in \newref{Table}{tab:notation}.

\begin{table}[!ht]\centering
\caption{List of notations}
\label{tab:notation}
\begin{tabular}{ll}\hline
 $G (V, E)$ & Transportation network graph \\
 $R$ & OD pairs set \\
 $V$ & Set of nodes in $G$ with size of $\left| V \right| = N$  \\
 $E$ & Set of links in $G$ with size of $\left| E \right| = L$ \\
 $Z$ & Set of classes in $G$ with size of $\left| Z \right| = n $ \\
 $m$ & Number of link features \\
  $n$ & Number of classes in the network \\
 $k$ & Number of feasible paths for each OD pair\\
$a$ & Input tensor dimension \\
$d$ & Embedding size \\
 $\X \in \mathbb{R}^{N^2}$ & OD demand matrix \\
 $\P \in \mathbb{R}^{N^2}$ &  Feasible path set \\
 $\tilde{\I} \in \mathbb{R}^{N^2\times a}$ & Normalized input tensor\\
 $\tilde{\F} \in \mathbb{R}^{N^2\times k}$ & Normalized path flow distribution tensor\\
 $\F^* \in \mathbb{R}^{N^2\times k}$  & Optimal path flow distribution tensor \\
 $\widehat{\F}^* \in \mathbb{R}^{N^2\times k}$  & Predicted path flow distribution tensor\\
 $\W_Q, \W_K, \W_V$ & Learnable weight matrix \\
 $\B$ & Single-head attention output \\
 $\G$ & Multi-head attention output \\
 $\W_o, \W_1, \W_2, \W_3, \W_4$ & Parameters of the feed-forward layer \\
 $\bigoplus$ & Concatenation operator \\
 \hline
\end{tabular}
\end{table}

After stacking and normalizing these tensors, the resulting input tensor is denoted as $\tilde{\I}$. As shown in the \newref{Figure}{fig:model-flow}, this normalized tensor $\tilde{\I}$ is then fed into the Encoder block.
The path flow distribution $\F^*$ is what our model needs to learn and predict. $\F^*$ is normalized before being fed into the Decoder block. The normalized tensor is $\tilde{\F}$.

\subsubsection{Attention Mechanism} \label{subsec4.2}
As the initial step of feature extraction, we employ a key-query-value mechanism on the input tensor in the Encoder block, and on the output tensor (path flow distribution tensor) in Decoder block, in addition to the weighted sum operation, to capture the importance of each OD pair (see \newref{Figure}{fig:model-flow}). 
In single-head attention, OD pair features are transformed into key, query, and value matrices, denoted by $\K, \Q, \V$ respectively, via the following linear transformations:
\begin{equation}\label{eq:kqv}
\centering
   \mathbf{Q, K, V} = [\tilde{\I}\W_Q, \tilde{\I}\W_K, \tilde{\I}\W_V], \quad \forall \W_Q, \W_K, \W_V \in \mathbb{R}^{a \times d}
\end{equation}

The value matrix captures the learned representations and importance scores for each pair, the query matrix $\Q$ represents the target pair's characteristics, and the key matrix $\K$ contains the features of other pairs. Attention scores between the query vector and each key are calculated using the scaled dot-product, which is then normalized by a softmax function to ensure the sum of the weights equals to one. These attention scores are subsequently used in a weighted sum operation with the value vector $\V$(\newref{Equation}{eq:attn}), allowing the model to focus on the most relevant output vectors for a given input vector.
\begin{equation}\label{eq:attn}
Attention \left(\Q, \K, \V\right) = \B = softmax \left(\dfrac{\Q\K^T}{\sqrt{d\K}}\right)\V,
\end{equation}
where $d\K$ is the last dimension of the key vector $\K$, $\B \in \mathbb{R}^{N^2 \times d}$.

For multi-head attention, these steps are repeated $h$ times to produce multiple attention outputs, where $h$ is the number of heads. 
The attention output tensors are concatenated along the last dimension, and a learnable weight matrix $\W_o$ is applied through \newref{Equation}{eq:mha} to return the final output tensor $\G$ with the same shape as the input tensor $\I$.
\begin{equation}\label{eq:mha}
Multi-head \left(\Q, \K, \V\right) = \G = \bigoplus_{\substack{i=0}}^{h-1} \B_i \W_o, \quad \forall \W_o \in \mathbb{R}^{dh \times a}
\end{equation}

\begin{figure}[!b]
    \centering
    \includegraphics[width=1\textwidth]{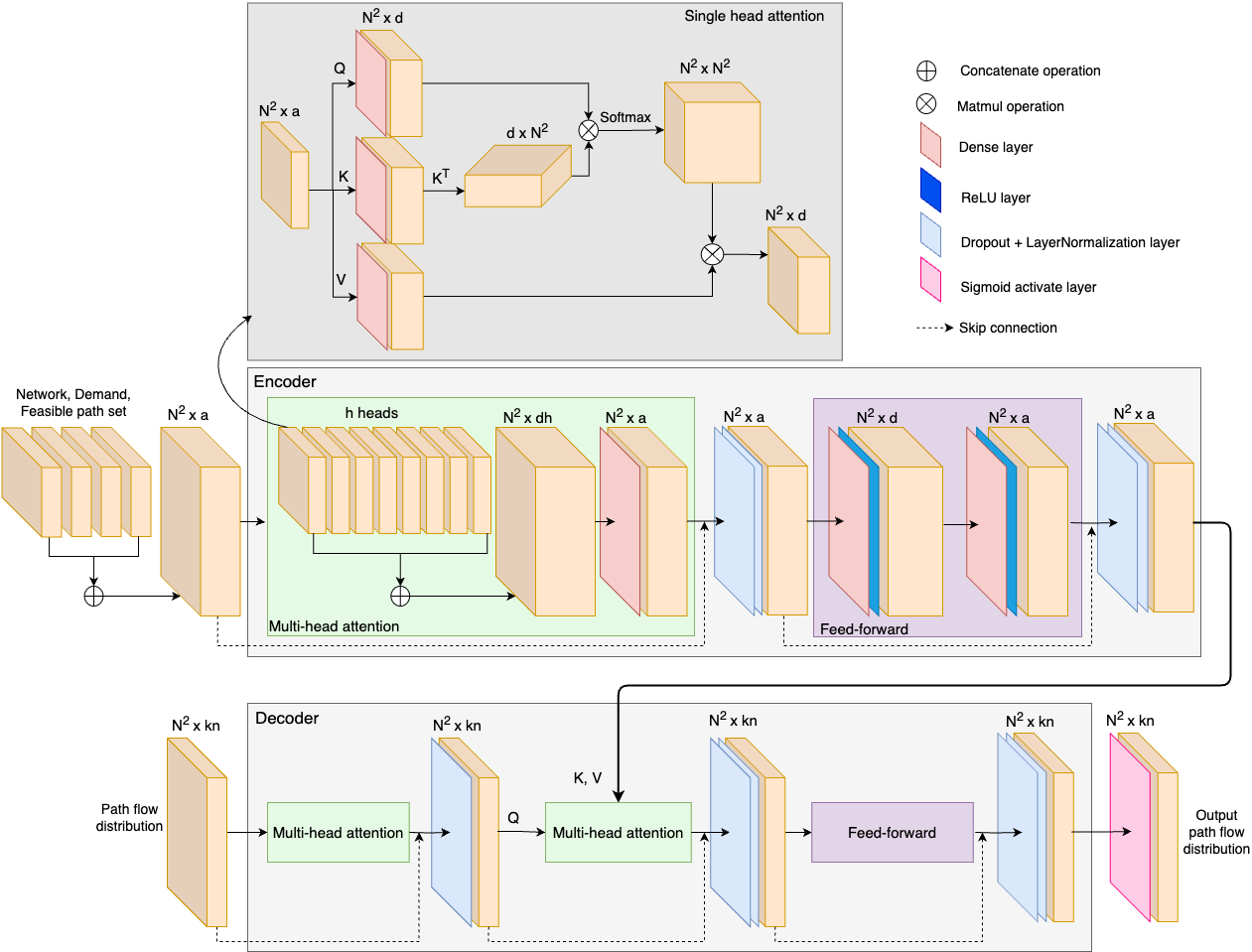}
    \caption{Model detail architecture}
    \label{fig:transformer-detail}
\end{figure}

\newref{Figure}{fig:transformer-detail} provides a comprehensive illustration of the multi-head attention mechanism's operational details.

\subsubsection{Encoder Layers} \label{subsec4.3}
The Encoder layer consists of two main components: a multi-head attention mechanism and a feed-forward neural network, both followed by residual connections and two layers of normalization. The detailed steps of each encoder layer include adding the output of multi-head attention to the input tensor, called a residual block. This residual connection helps mitigate the vanishing and exploding gradient problems, which are common issues in deep networks. It leads to faster convergence during training because it allows the model to learn identity mappings more easily \cite{He_2016_CVPR}. We perform dropout, merge in residual connections, and perform two-layer normalization to stabilize the training process (\newref{Equation}{eq:9}). 
    \begin{equation} \label{eq:9}
\E^{'} = LayerNorm\left(LayerNorm\left(\tilde{\I} + Dropout\left(\G\right)\right)\right)
\end{equation}
The output is then passed through a feed-forward neural network with two hidden layers, using ReLU activation functions. 
The feed-forward layer's output is added to the output of the previous step (the second residual block), followed by another two layers of normalization, \newref{Equation}{eq:10}, to stabilize the training:
\begin{equation} \label{eq:10}
\E = LayerNorm\left(LayerNorm\left(\E^{'} + Dropout\left(\W_1ReLU\left(\W_2\E^{'}\right)\right)\right)\right),
\end{equation}
\noindent where $\W_1$ and $\W_2$ are shared across all the positions in a layer.

Each Encoder layer processes the input tensor sequentially, capturing hierarchical relationships within the network.
The encoder block comprises multiple encoder layers stacked together. 
The output from one encoder layer serves as the input to the next, enabling the model to learn increasingly abstract representations of the input data.

\subsubsection{Decoder Layers} \label{subsec4.4}
The Decoder layer generates the output path flow distribution by attending to both the previous Decoder layer's output and the Encoder's output. 
Similar to the Encoder, the Decoder starts with multi-head attention on its input. 
The output of the first multi-head attention block on path flow tensor is denoted by $\G_1$. 
We also apply dropout, residual connection and layer normalization on this output (\newref{Equation}{eq:11}):
\begin{equation} \label{eq:11}
\D^{'} = LayerNorm\left(\tilde{\F}+ Dropout\left(\G_1\right)\right)
\end{equation}
The decoder then applies multi-head self-attention using the encoder's output $\E$ as the key and value, and the decoder's multi-head attention output $\D'$ as the query. This allows the decoder to focus on relevant encoder information. 
 $\G_2$ denotes the output of this second multi-head attention. As shown in \newref{Figure}{fig:model-flow}, the second residual connection and 2 layers of normalization are repeated before going to the feed-forward network (\newref{Equation}{eq:12}). 
 In \newref{Equation}{eq:13}, the result of the second residual connection is fed to a feed-forward layer, then the third residual connection is applied before the last 2 normalization layers:
\begin{equation}\label{eq:12}
\D^{''} = LayerNorm\left(LayerNorm\left(\D^{'} + Dropout\left(\G_2\right)\right)\right)
\end{equation}
\begin{equation}\label{eq:13}
\D = LayerNorm\left(LayerNorm\left(\D^{''} + Dropout\left(\W_3ReLU\left(\W_4\D^{''}\right)\right)\right)\right),
\end{equation}
where $\W_3$ and $\W_4$ are shared across all the positions in a layer.
The decoder block is composed of multiple decoder layers. The final output of the decoder is a tensor of size $N^2\times kn$, after a sigmoid activate function, we get a tensor representing the normalized path flow distribution for $k$ feasible paths and $c$ classes of each OD pair:
\begin{equation}\label{eq:14}
\widehat{\F}^* = sigmoid(\D)
\end{equation}

The Encoder and Decoder are two main blocks of this model in order to accomplish the prediction task (\newref{Equation}{eq:14}). In the next section, we apply the proposed architecture to two test cases with different characteristics. Note that given predicted path flows $\hat f^{r,z}_p$, we report:
\begin{align}
\varepsilon_{\text{OD}} &:= \frac{1}{|R|\,n} \sum_{r\in R}\sum_{z\in Z} \left| \sum_{p\in \P_r} \hat f^{r,z}_p - x_r^z \right|, \label{eq:odgap}\\
\varepsilon_{\text{link}} &:= \frac{1}{|E|} \sum_{e\in E} \left| \sum_{r\in R}\sum_{z\in Z}\sum_{p\ni e} \hat f^{r,z}_p - \hat v^{\,z}_e \right|, \label{eq:linkgap}\\
\phi_{\text{KKT}} &:= \frac{1}{|R|\,n} \sum_{r\in R}\sum_{z\in Z}\sum_{p\in \P_r} \hat f^{r,z}_p \,\max\!\big(0,\ c^{r,z}_p - u_r^z \big), \label{eq:kktr}
\end{align}
which quantify OD-conservation error, path-to-link aggregation consistency, and complementarity residuals, respectively. The model is penalized during training to ensure that these errors remain negligibly small (ideally near zero).


\section{Numerical Experiments}\label{sec:numerical}
Two numerical experiments are conducted to evaluate the accuracy, efficiency, and generalization capability of the proposed framework. In Section \ref{synthetic_data}, We first experiment with a synthetic Manhattan-like network to assess the basic effectiveness of the model. We evaluate the model’s performance under conditions of incomplete OD demand. The second experiment in Section \ref{urban_networks} is on urban transportation networks (Sioux Falls and Eastern-Massachusetts network) with several `what-if' scenarios to examine robustness against various realistic network perturbation scenarios. The implementation details and datasets used in these experiments are available at \url{https://github.com/sulthanashams/DL_TAP/tree/main}. The details of the experiments will be explained in the following sections.

\subsection{Training setup}
In this study, we compared the performance of various gradient descent optimization algorithms, including Stochastic Gradient Descent (SGD), RMSprop, Adagrad, and Adam. 
Our findings indicate that the Adam algorithm is the most effective for this research, offering superior accuracy and faster convergence. 
The model was trained using the Mean Squared Error (MSE) loss function, with the hyperparameters detailed in \newref{Table}{tab:2}. 
This set of hyperparameters has been validated as particularly effective when employing Transformers on large generated datasets \citep{charton2021}.
\begin{table}[!ht]
\caption{Hyperparameters used for model training}
\label{tab:2}
\centering
\begin{tabular}{lccc}
\hline
\textbf{Parameter} & \textbf{Value} & \textbf{Parameter} & \textbf{Value} \\ \hline
Batch size         & 64             & Attention heads    & 8              \\ 
Dimension          & 128            & Epochs             & 100            \\ 
Encoder layers     & 8              & Learning rate      & 0.001          \\ 
Decoder layer      & 1              & Dropout rate            & 0.1            \\ \hline
\end{tabular}
\end{table}

\subsection{Model evaluation indicator}
To evaluate the performance of the model, four measures are considered, one based on the UE condition and three classical measures. Based on the first condition of the UE principle (\newref{Equation}{eq:1}), we calculate the difference of average delay (AD) of the network between the predicted flow and the solution obtained from the optimizer as follows: 
\begin{equation}\label{Equation16}   \text{AD}^z = \frac{\displaystyle\sum\limits_{\forall{r}}\sum\limits_{\forall{p}}\hat{f}^{r, z}_p(c^{r, z}_p - u^z_r)}
    {\displaystyle\sum\limits_{\forall{r}}x^z_r} \quad \forall z \in Z
\end{equation}
In other words, a smaller difference, ideally close to zero, indicates a better-performing model. {To address the potential scaling effects of the denominator, we additionally examined the distribution of magnitude of the excess cost relative to mean travel times, per-OD pair in \ref{AppA}.}

Other metrics to evaluate model performance include: mean absolute error (MAE), mean absolute percentage error (MAPE):
\newcommand{\n}{n}
\begin{equation}
\text{MAE}^z = \frac{1}{y} \sum_{\forall r} \sum_{\forall p}\left|{f^{r, z}_p}^* - \hat{f}^{r, z}_p \right| \quad \forall z \in Z
\end{equation}
\begin{equation}
\text{MAPE}^z = \frac{1}{y} \sum_{\forall r} \sum_{\forall p} \left| \frac{{f^{r, z}_p}^* - \hat{f}^{r, z}_p}{{f^{r, z}_p}^*} \right| \times 100\% \quad \forall z \in Z
\end{equation}
where ${f^{r, z}_p}^*$ and $\hat{f}^{r, z}_p$ respectively represent the "ground truth" (optimal value) and predicted value of path flow, $y$ represents the total number of feasible paths of all OD pairs. It is important to note that while link flows may be unique, path flows are not necessarily so. In particular, when the network is homogeneous, it is not possible to establish a definitive "ground truth" for the distribution of path flows. In such cases, the only means of validating the results is through the use of quality indicators.

\subsection{Manhattan-like road network}\label{synthetic_data}
In this section, we examine the generalization capability of the proposed model to a Manhattan-like network. We consider a realistic scenario where the regional OD demand values are incomplete. In this way, the model is expected to learn the inherent patterns and structures of the transportation network, enabling reliable flow prediction even when parts of the OD demand information are missing, as commonly observed in real-world data collection.
\subsubsection{Data description}

As part of our experimental setup, we generate a grid network with 25 nodes and 80 links. 
The detailed information of the link, including capacity, length, and free flow travel time, is randomly generated following uniform distribution (see \newref{Table}{tab:syn-net-detail}). 
For the 1st analysis, we generate data and train the model with a single class (car). \newref{Figure}{fig:syn_net} presents the heatmap of link capacity for each link in the bi-directional graph.
\begin{table}[!ht]
\caption{Manhattan-like road network characteristics}
\label{tab:syn-net-detail}
\centering
\begin{tabular}{lcc}
\hline
\textbf{Measure}      & \textbf{Min value} & \textbf{Max value} \\ \hline
Length (km)           & 20                 & 40                 \\ 
Capacity (veh)        & 1000               & 2000               \\ 
Free flow time (hour) & 0.5                & 1                  \\ 
Demand                & 50                & 500               \\ \hline
\end{tabular}
\end{table}
\begin{figure}[!h]
    \centering
    \includegraphics[width=0.8\textwidth]{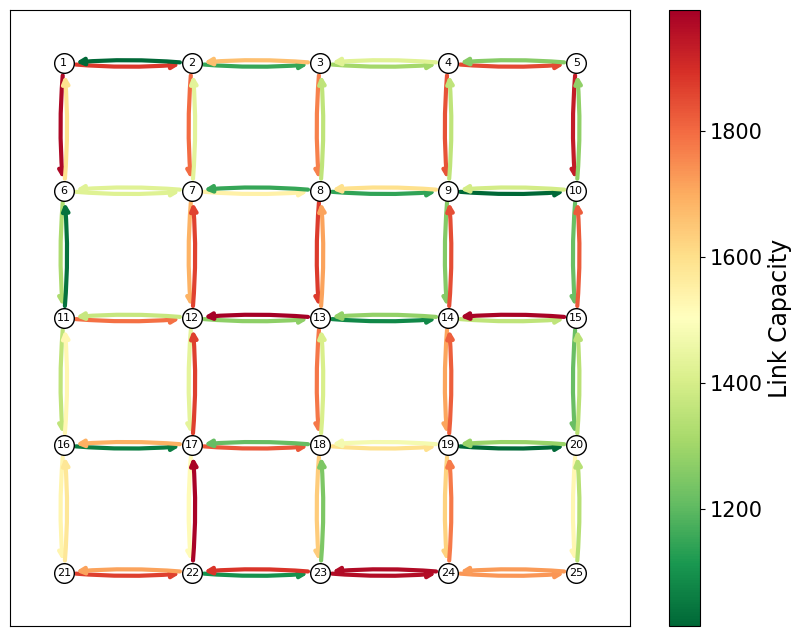}
    \caption{Sampled generalized synthetic networks with 25 nodes and 80 links}
    \label{fig:syn_net}
\end{figure}

For the experiment, incomplete OD demand values are achieved by randomly removing a ratio of OD pairs. 
Let $s$ denotes the number of complete OD pairs $(v1, v2)$, with $v1 \neq v2$, $s = N^2 - N$, $j$ denotes the missing percentage.
Two missing ratios are considered in this experiment: $j = 30\%$ and $j = 40\%$.
With this experiment, we can perform `what-if' analyses where the demand changes.
To create the training, testing, and validation datasets, we generate 4000 OD demand matrices at each missing ratio for the graph ($s-j$ OD pairs). 
This approach enables the model to learn the inherent patterns and structures of the transportation network, even when some demand information is missing. {For each incomplete-OD scenario, the UE reference solution is computed using the same OD matrix with missing entries (rather than the corresponding complete-demand matrix), and this scenario-consistent UE solution is used as the ground truth for training and evaluation.}

In this study, we set $k=3$ for shortest paths exploration per each OD pair. If an OD pair has fewer than three feasible paths, the path set and path flow are padded with zeros to reach a dimension of three.
The optimal solutions for UE of 4000 demand scenarios at each missing ratio are then obtained by utilizing the Gurobi optimizer.
The solution has some delays due to the linearization. Because our aim is to learn and estimate the solution accurately, we will accept this error and assume that the network has reached an equilibrium state. 
The training, validation, and testing sets are then divided into 70\%, 20\%, and 10\%, respectively. 
We train the model with the missing rate $j = 30\%$, then test the model at two missing rate $j = 30\%$ and $j = 40\%$. Thus, a pre-trained model on $j=30\%$ was directly applied to predict outcomes under $j=40\%$ absence, providing a more practical approach for real-world use.

\subsubsection{Numerical results}
The training history of the model is shown in \newref{Figure}{fig:syn_training}. The model converges smoothly after 80 epochs with few fluctuations. 
After predicting path flow distribution, the link flows are then aggregated to calculate the error at both path and link levels.
\newref{Table}{tab:syn_result} summarizes the prediction performance of the proposed model under different OD demand missing ratios. 
When the missing ratio increases, the proposed model maintains a relatively low MAPE. For instance, in the scenario where 30\% of OD demand is missing, the MAPEs for predicted link flow and path flow are 2.56\% and 5.50\%, respectively. 
When the missing ratio increases to 40\%, the model provides link flow and path flow MAPEs of 4.35\% and 7.17\%, respectively. The average network delay resulting from the predicted path flow remains under 2\% in both scenarios, though it does increase as the missing ratio of OD pairs rises. Representing only a slight increase compared to the $30\%$ missing ratio, the model was able to generalize reasonably well to a $10\%$ higher OD missing case without retraining. 

\begin{figure}[!ht]
    \centering
    \includegraphics[width=1\textwidth]{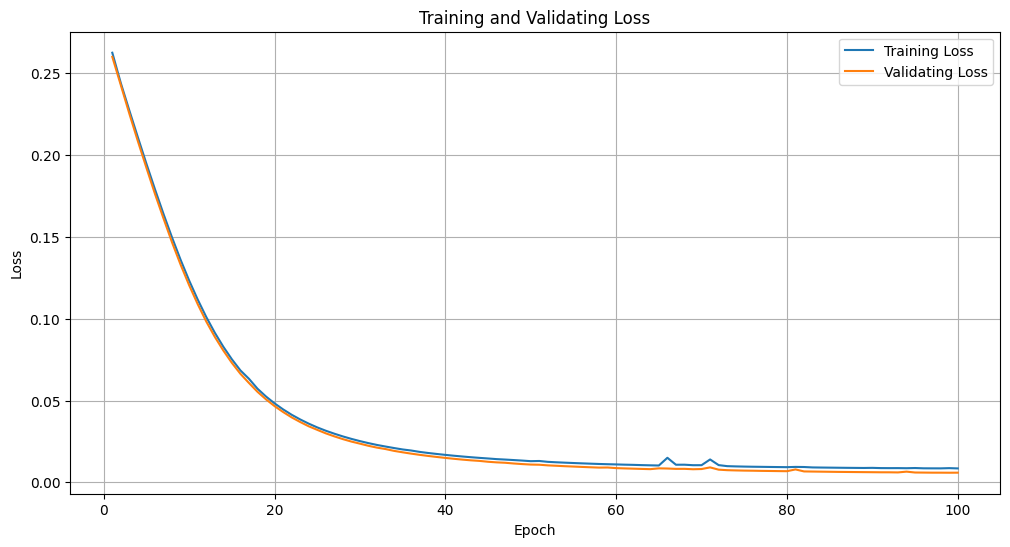}
    \caption{Training history of model with synthetic network}
    \label{fig:syn_training}
\end{figure}

\begin{table}[!ht]
\centering
\caption{Model performance under different OD demand missing ratios}
\label{tab:syn_result}
\begin{tabular}{lcc}
\hline
\textbf{Indicator} & \textbf{Missing ratio = 30\%} & \textbf{Missing ratio = 40\%} \\ \hline
Path flow MAE & 10.95 & 18.54 \\ 
Path flow MAPE (\%) & 4.50 & 5.57 \\ 
Predicted Delay (\%) & 1.22 & 1.38 \\ 
UE Solution Delay (\%) & 0.0001 & 0.0001 \\ \hline
\end{tabular}
\end{table}

In addition to the prediction accuracy, the training time is also an important factor in evaluating the efficiency and practicality of machine learning models. {The training time is 72 minutes, while generating the optimization-based reference dataset (i.e., solving the UE problem for all 4000 demand scenarios) required 210 minutes in total.} 
The prediction time for a single OD demand matrix is 0.001 seconds, it is 5000 times faster than using the Gurobi optimizer (5 - 7 seconds per OD matrix).
These results highlight the computational efficiency of the proposed model in addressing traffic assignment problems.

\subsection{Validation on urban transportation network}\label{urban_networks}
Having observed the model's strong performance with the synthetic Manhattan-like network, we evaluate its effectiveness on an urban transportation network. This step allows us to assess its effectiveness under more realistic conditions, where traffic systems exhibit greater heterogeneity and complexity. Starting from the baseline scenario of incomplete OD data, we introduce additional training cases that incorporate varying levels of missing information and structural constraints. Corresponding testing sets are then defined for each scenario to systematically examine the model’s robustness and adaptability across different urban traffic settings.

\noindent Note that the OD demand values and the number of OD demand matrices remain fixed for the analysis. We explore the following scenarios on the urban transportation network: 

    \subsection*{ Scenario 1: Both demand and network supply change}
    We consider cases with varying link missing ratios alongside partially missing OD demand data. This setup reflects realistic conditions of data unavailability in urban transportation systems, where incomplete or unreliable measurements on both demand and supply sides are common. The scenario is designed to evaluate the model’s ability to perform reliably under such simultaneous data limitations. 
    
    \subsection*{ Scenario 2:  Some links in the network are disabled randomly} 
   Specifically, we evaluate the model’s performance when 2–3 links are randomly removed from operation. This setup reflects realistic structural disruptions in urban transportation systems, such as road closures due to accidents, construction works, or policy-driven restrictions. Importantly, these cases are tested without retraining the model on the newly disabled links, to assess its ability to generalize to minor unforeseen disruptions in network structure.

   { \subsection*{Scenario 3: Generalisation to Topological Variations}
    In this scenario, we generate a set of synthetic networks by modifying the original SiouxFalls urban network used in the previous experiments. This setup simulates structural changes such as urban expansion or the introduction of new routes resulting from the closure or redevelopment of existing areas. The model is trained on the original SiouxFalls network and tested on these topological variations to assess its generalization capability to unseen network structures.}
    
    \subsection*{Scenario 4: There are multiple classes in the network (car and truck).} 
    This scenario introduces heterogeneity in demand patterns and traffic dynamics, thereby testing the model’s ability to account for varying impacts of different vehicle types on network flows.

    \subsection*{ Scenario 5: When the network scale increases (large-scale network).} 
    Finally, we test the model on two larger-scale urban networks to examine its scalability and robustness in more complex settings. Compared to the synthetic and medium-sized urban networks, this scenario involves a greater number of OD pairs, links, and potential sources of congestion. It allows us to assess the model’s efficiency and effectiveness when applied to realistic, large-scale transportation systems.

In the following section, we describe the network and experimental setup for each scenario, including the urban network characteristics, OD demand inputs, and the design of training and testing datasets.

\subsubsection{Data description and Experiment Set-up}
In scenarios 1, 2, 3 and 4 we utilize Sioux Falls network. The information about this network is sourced from \cite{Bar-Gera}. This network comprises 24 nodes and 76 links. \newref{Table}{siouxfalls} and \newref{Figure}{fig:demand_hist} illustrates the network's characteristics. 

\begin{table}[h!]
\centering
\caption{SiouxFalls network details}
\label{siouxfalls}
\begin{tabular}{lcc}
\hline
\textbf{Detail}                 & \textbf{Min value} & \textbf{Max value} \\ \hline
Length (km)                     & 2                  & 10                 \\
Capacity (vehicle)              & 4823               & 25900              \\ 
Free flow travel time (minutes)  & 2                  & 10                 \\ 
Demand                          & 100                 & 4000               \\ \hline
\end{tabular}
\end{table}
\begin{figure}[h!]
    \centering
    \includegraphics[width=1\textwidth]{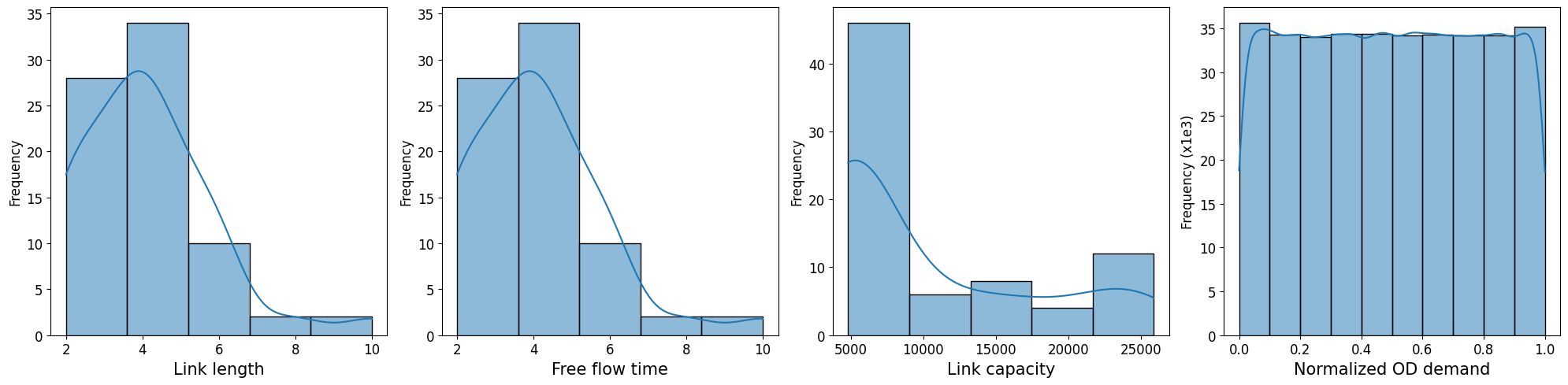}
    \caption{Histogram of Sioux Falls network characteristics with trend line }
    \label{fig:demand_hist}
\end{figure}

The original dataset comprises link lengths ranging from 2 to 10 kilometers and free-flow travel times spanning 2 to 10 minutes. The histograms of these two variables exhibit a right-skewed distribution, indicating that the majority of links are shorter in length and have shorter free-flow travel times. 
Most links have capacities between 5,000 and 10,000, while links with capacities exceeding 20,000 constitute a small percentage of the dataset \cite{Bar-Gera}.

In the first scenario of this experiment, to conduct a comprehensive `what-if' analysis, we combine both the change in demand (incomplete demand) and in network supply (where some links are missing). We use a single-class network in this scenario for simplicity. Let $L$ denotes the number of links, and $j$ denotes the percentage of missing links.
We consider that the regional OD demand is missed by 30\%, and two link missing ratios are examined: $j = 5\%$ and $j = 10\%$. 4000 different OD demand matrices at this demand missing ratio are generated. The demands follow uniform distribution (\newref{Figure}{fig:demand_hist}).
The optimal path flow is then obtained by the optimizer, presented in the previous section.
\newref{Figure}{fig:pathFlow_hist} and \newref{Figure}{fig:boxplot_linkflow} depict the changes in the optimal path flow and link flow distribution under different link missing ratios. 
Most of the flows are distributed to the first path, since normally this is the shortest path. As the missing ratio increases from 0\% to 5\%, the flow will be distributed to other alternative paths (the second and third paths).
The changes in path flow lead to the changes in link flow. 
In the \newref{Figure}{fig:boxplot_linkflow}, we note that when the network misses 10\% of links, the flows of links 1-6 and 24 increase (red boxes), while the flows of links 48 and 53-60 decrease (green boxes). 

\begin{figure}[h!]
    \centering
    \includegraphics[width=1\textwidth]{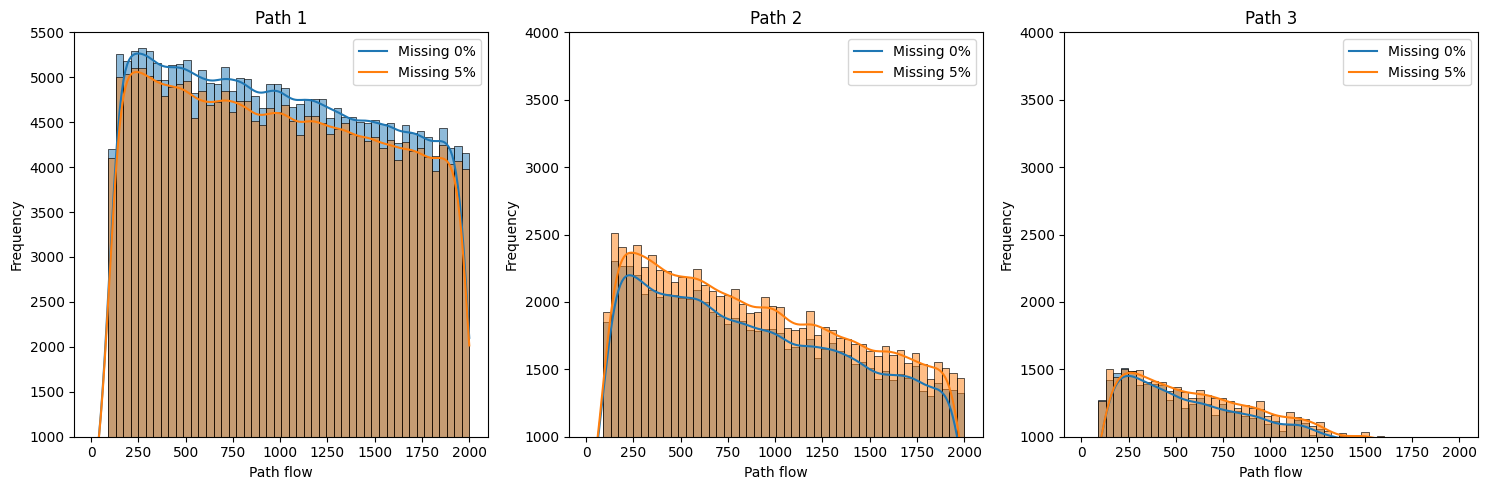}
    \caption{Histogram of optimal path flow distribution of difference link-missing ratios}
    \label{fig:pathFlow_hist}
\end{figure}
\begin{figure}[h!]
    \centering
    \includegraphics[width=0.9\textwidth, height=0.5\textheight]{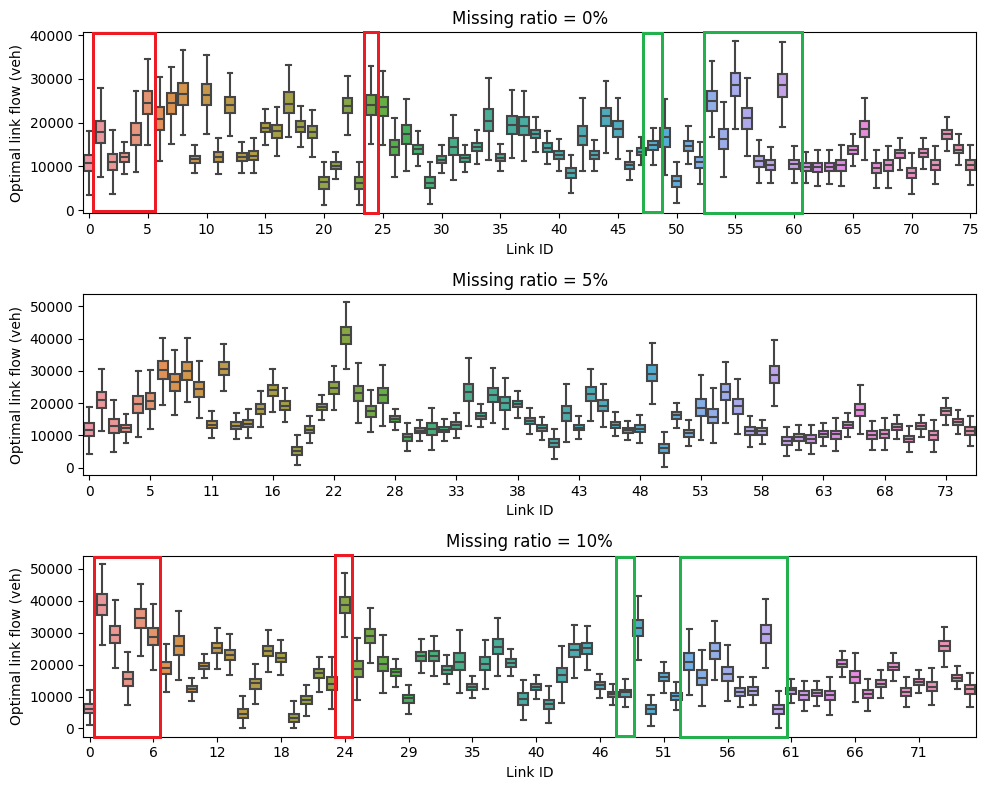}
    \caption{Boxplot of optimal link flow distribution for different link-missing ratios}
    \label{fig:boxplot_linkflow}
\end{figure}

Having generated the optimal dataset to serve as the ground truth for performance evaluation, we proceed to train the model to obtain its predictions. The training settings and hyperparameters are kept identical to those used in the previous experiment. The OD demand values and the number of OD matrices are kept constant across all experiments, with $30\%$ of OD entries missing and a total of $4000$ OD matrices generated. Scenarios 1-3 utilize the Sioux Falls network, which consists of 24 nodes and 76 links, {while Scenario 4, designed to test scalability, employs the larger \textit{Eastern Massachusetts} (EMA) network, which includes 74 nodes and 258 links and the \textit{Anaheim} Network, which includes 38 zones covering 416 nodes and 914 links. The experimental setup for each scenario is described as follows.}

In \textit{Scenario~1}, the model is trained and tested on a network with $L-j$ links, where $j$ represents the proportion of removed links. 

In \textit{Scenario~2}, the model is first trained on the full network containing $L$ links, and then evaluated on networks where two and three links are randomly removed ($L-2$ and $L-3$, respectively). These scenarios use a single-class network.

{In \textit{Scenario 3},  we generate a set of synthetic networks by modifying the original SiouxFalls urban network used in the previous experiments. Specifically, 20 unique topological variants are created for three configurations of link-change fractions: 25\%, 50\%, and 75\%. Each variant is formed by randomly adding and removing links, with the feature values of new links computed as the mean of their neighbouring links to ensure realistic network characteristics. \newref{Figure}{fig:siouxfalls_variation} illustrates three representative randomly generated network variations corresponding to each case. The model is trained on the original SiouxFalls network and tested on these variants to assess its robustness under structural variations.}

In \textit{Scenario~4}, we extend the framework to a multi-class setting with $n = 2$ classes (cars and trucks). Truck demand is randomly generated as $50\%$ of the car demand, and the free-flow travel time for trucks is set to $1.5$ times that of cars. This scenario uses the full network with all $L$ links.

 {In \textit{Scenario 5}, we evaluate the scalability of the proposed framework using the larger EMA and Anaheim urban network. \newref{Table}{tab:network_details} provides detailed information about the EMA and Anaheim networks. For the Anaheim network, number of zones is not same as number of nodes, so for each path connecting the zones, the link features are computed as the average of the links comprising the connecting path. Additionally, to generate OD variations, we create multiple trip demand scenarios using a scaling factor on the base trip demand for Anaheim, resulting in 4000 OD variations. As this scenario focuses on assessing the model's performance with large-scale networks, we conduct a single experiment with the full network, where 50\% of the demand is missing.}

\begin{table}[H]
\centering
\caption{Network details for EMA and Anaheim}
\label{tab:network_details}

\begin{subtable}[t]{0.45\textwidth}
\centering
\caption{Eastern Massachusetts (EMA) network}
\begin{tabular}{lcc}
\hline
\textbf{Detail}                 & \textbf{Min value} & \textbf{Max value} \\ \hline
Length (miles)                  & 1.06               & 32.93              \\ 
Capacity (vehicles)             & 825                & 8352.01            \\ 
Free flow travel time (hours)   & 0.016              & 0.88               \\ 
Demand                          & 50                 & 500                \\ \hline
\end{tabular}
\end{subtable}
\hfill
\begin{subtable}[t]{0.45\textwidth}
\centering
\caption{Anaheim network}
\begin{tabular}{lcc}
\hline
\textbf{Detail}                 & \textbf{Min value} & \textbf{Max value} \\ \hline
Length (feet)                   & 264                & 9451               \\ 
Capacity (vehicles)             & 1800               & 12600              \\ 
Free flow travel time (mins)    & 0.05               & 3.57               \\ 
Demand                          & 1                  & 2106               \\ \hline
\end{tabular}
\end{subtable}

\end{table}

\subsubsection{Numerical results}
\noindent \textbf{Scenario 1:}


\newref{Table}{result} summarizes the prediction performance of the model under various scenarios of missing links. It is evident that the model can predict path flow distribution even with missing links, although it performs best when no links are missing. 
The predicted path flows result in a network delay of 0.79\% compared to the average path cost.
The error of predicted path flow is smaller than the error of link flow, which is aggregated from predicted path flow, with MAPE of 2.25\% for path flows and 2.61\% for link flows.



\begin{table}[h!]
\centering
\caption{Model performance with Sioux Falls network under different link missing ratios}
\label{result}
\resizebox{0.8\textwidth}{!}{
\begin{tabular}{lccc}
\hline
{\centering \textbf{Measure}} & \textbf{Missing ratio = 0\%} & \textbf{Missing ratio = 5\%} & \textbf{Missing ratio = 10\%} \\ \hline
Path flow MAE & 5.02 & 6.93 & 7.37 \\ 
Path flow MAPE (\%) & 2.25 & 2.98 & 3.07 \\ 
Prediction Delay (\%) & 0.79 & 1.52 & 1.69 \\ 
UE Solution Delay (\%) & 0.02 & 0.03 & 0.05 \\
\hline
\end{tabular}}
\end{table}

When the missing ratio increases to 10\%, the network delay percentage also increases from 0.79\% to 1.69\%. Consequently, the MAPEs for the predicted path flows and the link flows increase to 3.07\% and 9.06\%, respectively.
\newref{Figure}{fig:errors} illustrates the distribution of errors in link flow and the path flow in this scenario. The histogram is heavily skewed to the right, indicating that most errors are relatively small. For example, 95\% of the predicted link flows have an absolute error of less than 479, and 95\% of the predicted path flows have an absolute error under 23.

\begin{figure}[H]
    \centering
    \includegraphics[width=0.85\textwidth]{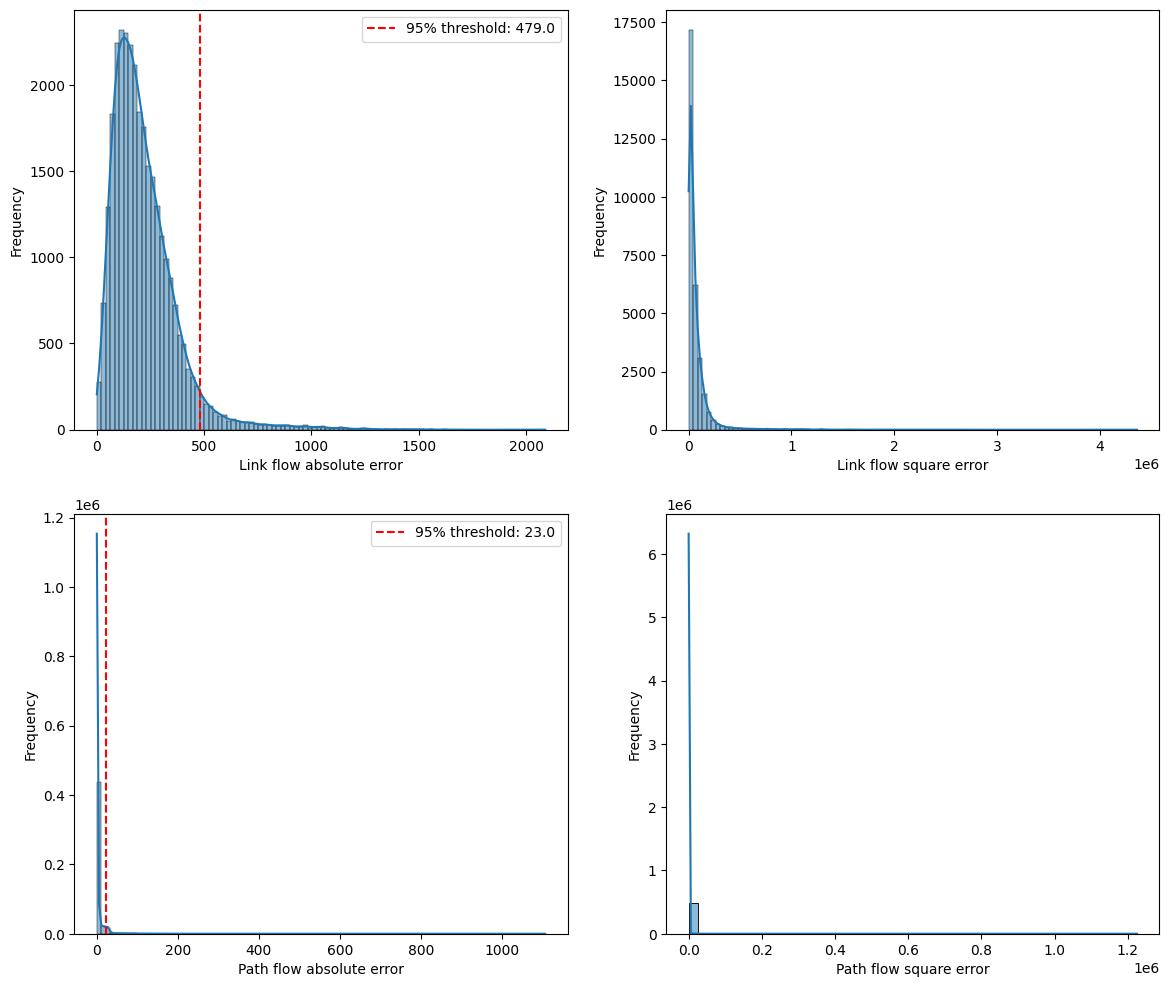}
    \caption{Histogram of link flow and path flow error}
    \label{fig:errors}
\end{figure}

{For the full Sioux Falls network with 0\% missing links, the path-level MAE of 5.02 corresponds to a link-level MAE of 256.27. This larger absolute error at the link level is expected because link flows are obtained by aggregating multiple path flows through the path-link incidence matrix. Therefore, even small errors at the path level may accumulate on links that are shared by many OD pairs, especially high-demand or frequently used links.}

{The error distribution in \newref{Figure}{fig:errors} should therefore be interpreted in light of the heterogeneous usage of links and paths across the network. Most predicted errors remain small, while larger errors are concentrated on links or paths carrying higher demand or serving as common segments for several OD pairs. From a practical perspective, the predicted path flows remain useful because they preserve route-level information and allow link flows to be recovered through aggregation. This is particularly important for planning and scenario analysis, where route-choice patterns and relative flow redistribution are often more informative than isolated link-level errors.}

\newref{Figure}{fig:heatmap} demonstrates the changes in the error of predicted flows at the link level as the link missing ratio increases. 
Note that when the link missing ratio increases to 10\%, the errors of the links between nodes 8-9 and 13-24 increase significantly due to the low capacity (5050 and 5091, respectively) and the high travel demand of all the paths that pass through these links. In particular, the frequency of usage of the link between nodes 8-9 and nodes 13-24 is 179 and 175, respectively, when the link missing ratio increases to 10\%. These are the highest frequencies values compared to other links.

{To highlight the proposed model’s accuracy and computational efficiency, we compare its performance against both traditional mathematical methods and a learning-based baseline. Classical traffic assignment algorithms, including MSA, Frank-Wolfe-type algorithms, and reduced-gradient approaches, have been widely studied and benchmarked in the TAP literature \citep{perederieieva2015framework,babazadeh2020reduced}. In this study, \newref{Table}{tab:heuristic_performance_comparison} presents a comparison with the Method of Successive Averages (MSA) \citep{MSA, bar-gera02}, the Gap Heuristic proposed by \cite{lu2009equivalent}, and a Multi-Layer Perceptron (MLP) baseline. The MLP is implemented as a fully connected neural network with four layers of sizes [256, 128, 64, 32]. All methods are evaluated on the full Sioux Falls network with 30\% missing OD demand.}

{The MSA solves the TAP by iteratively shifting a fraction of traffic from non-shortest to shortest paths, where at each iteration $i$ the swapped fraction is fixed as $\frac{1}{i+1}$. In the gap-based variant, this fraction is adaptively determined from the relative gap between path flows at each iteration, allowing larger shifts when the system is far from equilibrium and smaller adjustments as the gap decreases.
The high delays observed for MSA stem from its heuristic, iterative nature \citep{lu2009equivalent, ameli2020cross}. In early iterations, all demand is concentrated on a few shortest paths, and even after many iterations, Gap-based MSA remains far from equilibrium, resulting in inflated average delays.}

{Structurally, in the MLP model, the OD demand data and network data are flattened into a single vector as input. Thus, each OD pair is treated independently. This formulation lacks the ability to model interdependencies between ODs. As a result, it produces reasonable estimates at the individual OD level but fails to enforce network-wide consistency. In other words, the flattened representation of MLP does not explicitly encode the structured relationships among OD pairs, paths, and network topology. Instead, all dependencies must be learned implicitly through dense layers, without the benefit of attention-based interaction modeling. This structural disadvantage causes MLP to perform worse than MSA despite being trained on exact Gurobi UE solutions. In contrast, MSA inherently preserves these dependencies through its iterative link flow updates. In comparison, Gurobi-based optimization provides near-optimal solutions by explicitly solving the TAP as a convex optimization problem. However, as noted previously, optimization-based models require recalculating the entire network for even minor 'what-if` scenarios.}

{The Transformer model, trained on Gurobi-generated UE path flows, achieves high accuracy while offering substantially faster inference than both optimization-based frameworks and traditional heuristic approaches, as indicated by the computation time. It is important to note that the computation time reported for MSA corresponds to a solution that is still far from equilibrium, whereas, within the same time frame, the optimization-based framework attains near-optimal results. A trained Transformer model, on the other hand, predicts path flows roughly 1000 times faster than either approach for this network. The MLP predicts approximately 30 times faster than the Transformer due to its significantly simpler architecture.}

\begin{table}[h!]
\centering
\caption{Average delay (in percentage) and computation time of different methods on the original network (0\% missing links), sorted by performance}
\label{tab:heuristic_performance_comparison}
\begin{tabular}{lcc}
\hline
\textbf{Method} & \textbf{Average Delay (\%)} & \textbf{Computation Time (s)} \\ 
\hline
Gurobi Optimizer             & 0.03\% & 3 \\  
Transformer Prediction Model & 0.79\% & 0.003 \\  
Gap-Based MSA                & 4.56\% & 4.2 \\  
MLP                & 15.16\% & 0.0001 \\
\hline
\end{tabular}
\end{table}

\begin{figure}[H]
\centering
\begin{subfigure}{0.3\textwidth}
    \includegraphics[width=\textwidth]{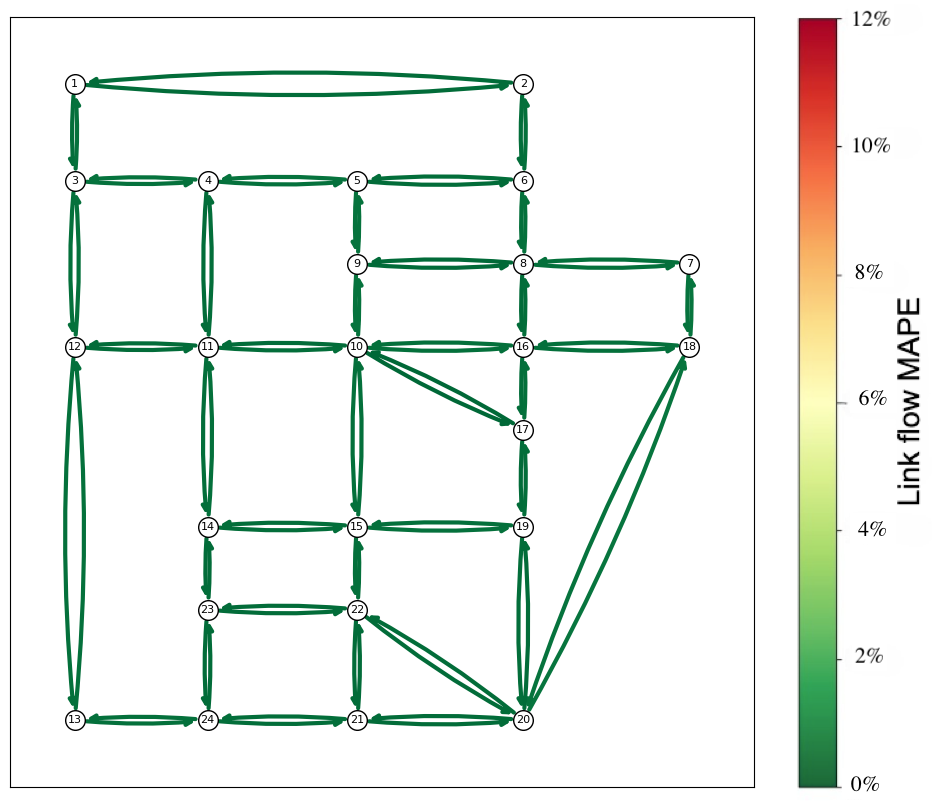}
    \caption{Missing ratio = 0\%}
    \label{fig:heat1}
\end{subfigure}
\hfill
\begin{subfigure}{0.3\textwidth}
    \includegraphics[width=\textwidth]{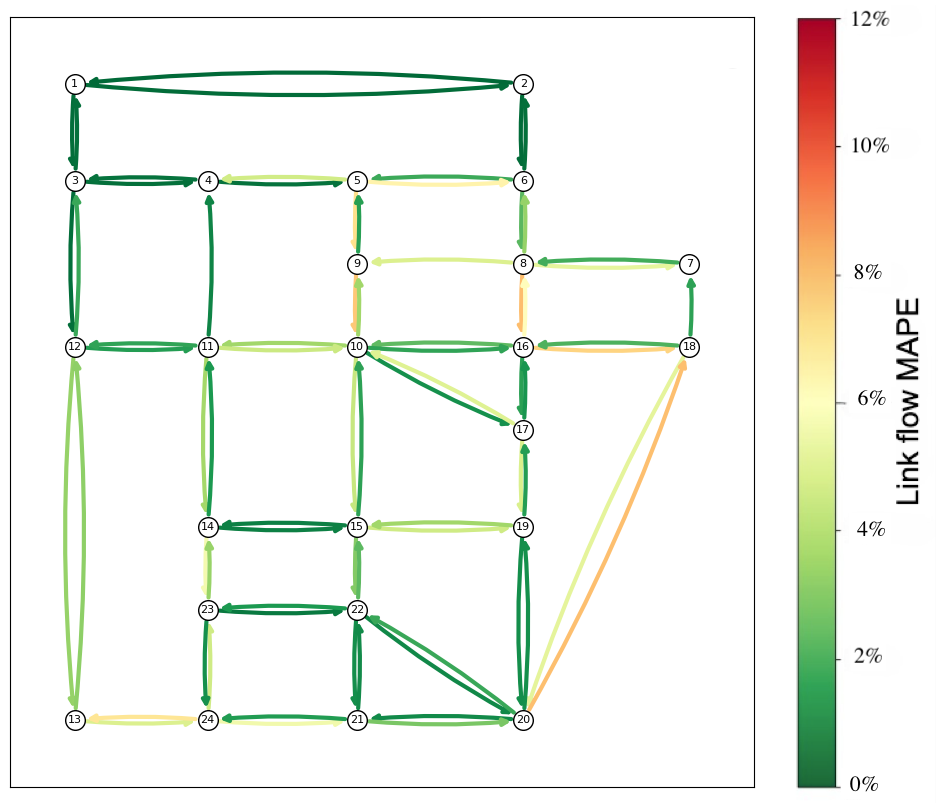}
    \caption{Missing ratio = 5\%}
    \label{fig:heat2}
\end{subfigure}
\hfill
\begin{subfigure}{0.3\textwidth}
    \includegraphics[width=\textwidth]{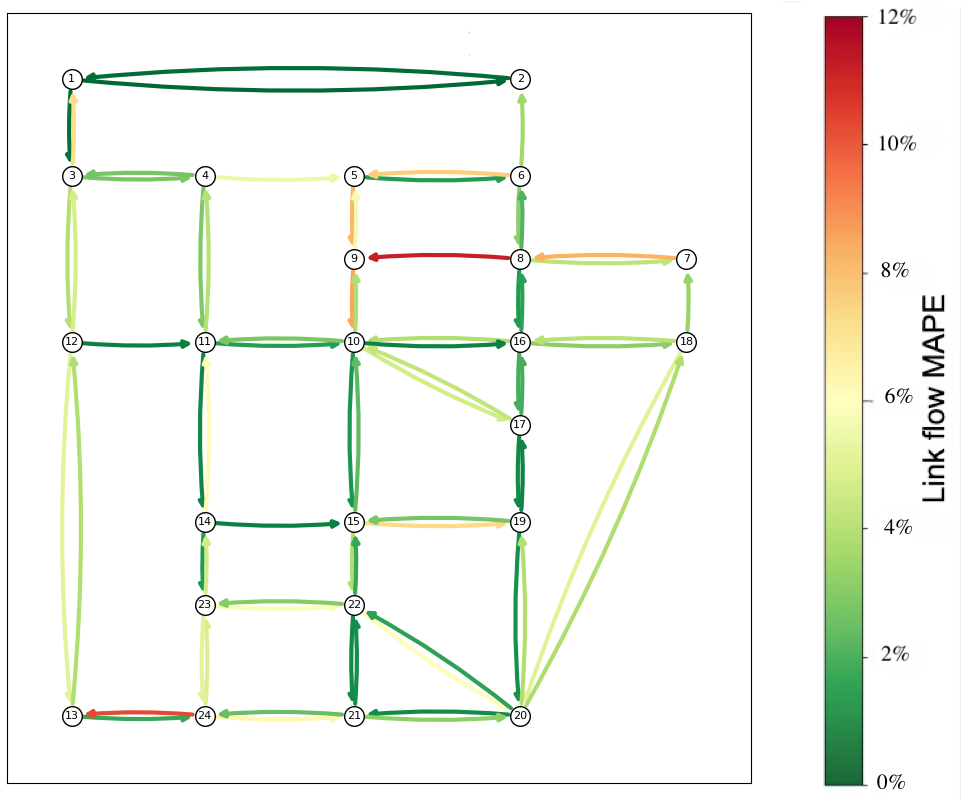}
    \caption{Missing ratio = 10\%}
    \label{fig:heat3}
\end{subfigure}
\caption{Heat map of link flow MAPE in different link missing ratios}
\label{fig:heatmap}
\end{figure}

\noindent \textbf{Scenario 2:}

\newref{Table}{result2} presents the model's performance when two and three links are randomly disabled without retraining. The results demonstrate that the proposed model is capable of accurately predicting the distribution of the path flow despite minor changes in the network supply, without requiring retraining. The predicted path flow exhibits an MAPE of less than 5\%, indicating high predictive accuracy with a small error. Furthermore, the average network delay associated with the predicted path flow remains low, at just 0.65\%. As the number of removed links increases from two to three, the MAPE of the predicted path flow rises from 3.61\% to 4.25\%.
The AD difference is minimal, showing the model's strong ability to predict travel times accurately. 

The result demonstrates the model's capability in conducting `what-if' analyses, particularly in scenarios where certain links are removed. For the model to function correctly, the only requirement is to ensure that at least one feasible path exists for each input OD pair.

\begin{table}[!ht]
\centering
\caption{Model performance with Sioux Falls network when randomly disabling 2-3 links without retraining}
\label{result2}
\begin{tabular}{lcc}
\hline
\textbf{Measure} & \textbf{2 links disable} & \textbf{3 links disable} \\ \hline
MAE & 8.07 & 8.16 \\ 
MAPE (\%) & 3.61 & 4.25 \\ 
Prediction Delay (\%) & 0.63 & 0.65 \\
UE Solution Delay (\%) & 0.04 & 0.03 \\
\hline
\end{tabular}
\end{table}

\vspace{5mm}

\noindent \textbf{Scenario 3:}

{\newref{Table}{tab:link_change_performance} presents the model's performance under varying levels of topological variation, where the model is trained solely on the original SiouxFalls urban network. The results show that, although MAE and MAPE increase as the degree of structural change grows, the predicted delay percentage remains relatively stable across all scenarios. This indicates that while path-flow predictions become less precise when the network structure deviates, the resulting deviations relative to overall network travel times are minimal, demonstrating robust predictive performance under network variations.}

\begin{figure}[H]
    \centering
    \begin{subfigure}[b]{0.32\textwidth}
        \centering
        \includegraphics[width=\textwidth]{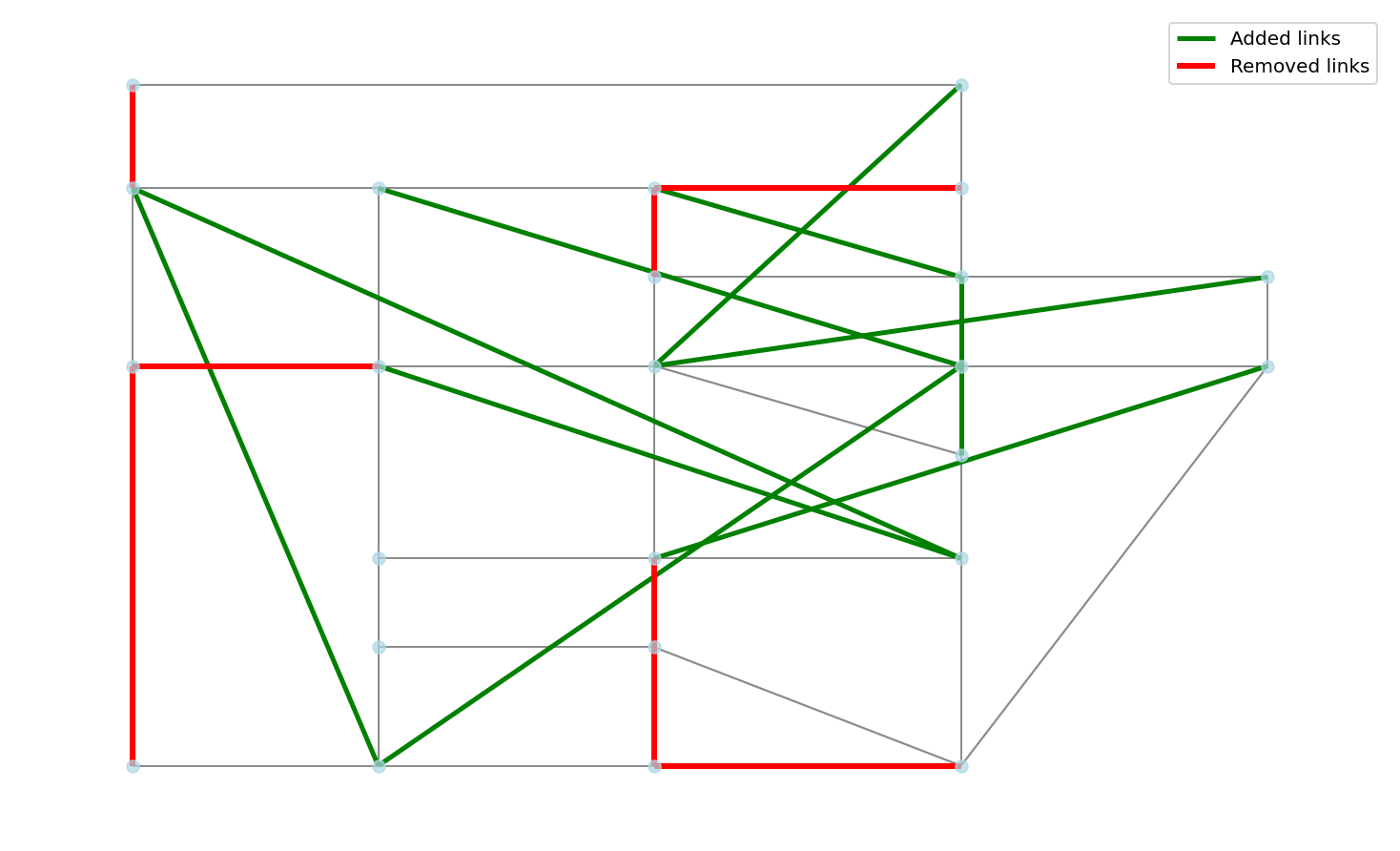} 
        \caption{25\% variation}
        \label{fig:sf_topology_25}
    \end{subfigure}
    \hfill
    \begin{subfigure}[b]{0.32\textwidth}
        \centering
        \includegraphics[width=\textwidth]{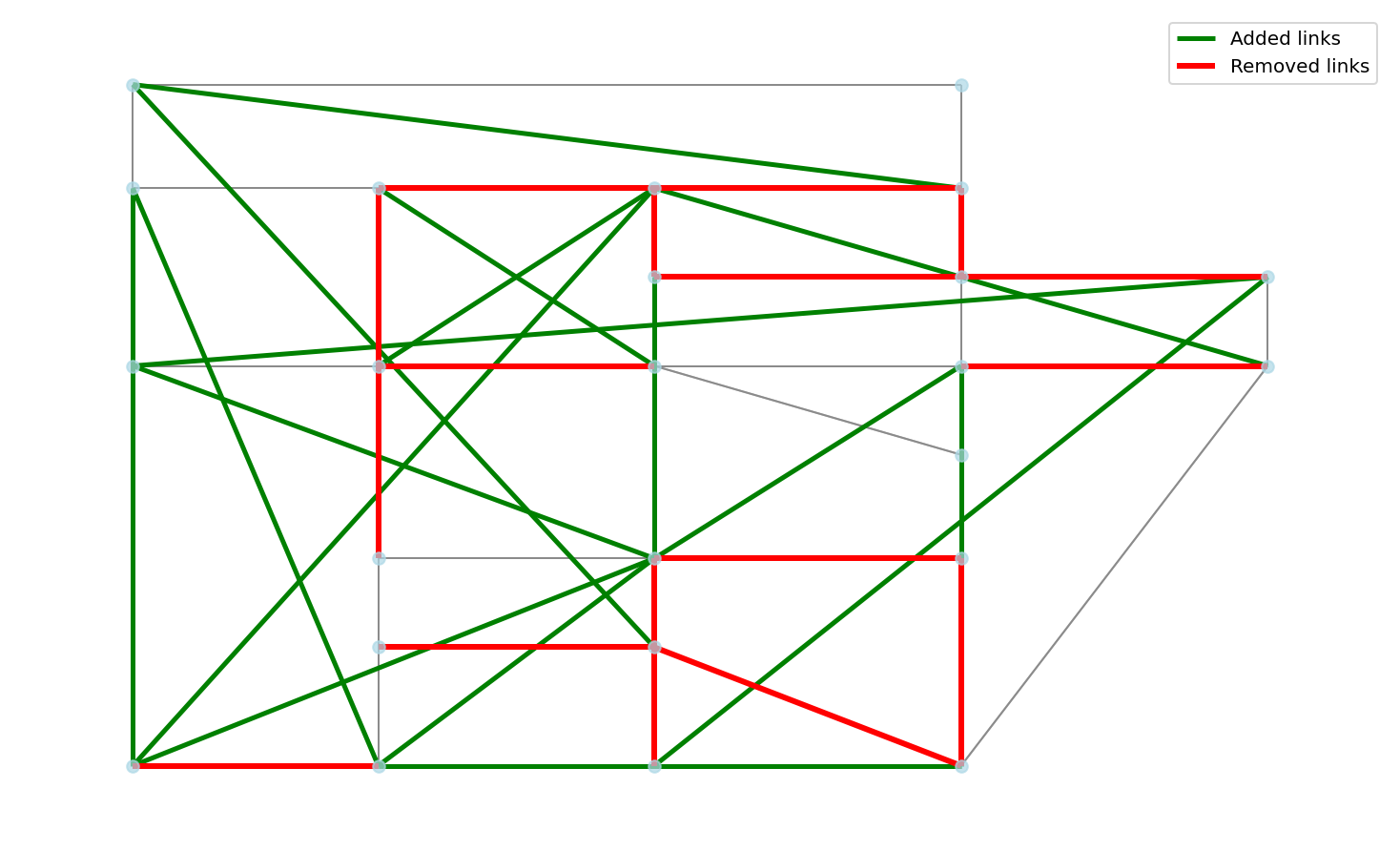} 
        \caption{50\% variation}
        \label{fig:sf_topology_50}
    \end{subfigure}
    \hfill
    \begin{subfigure}[b]{0.32\textwidth}
        \centering
        \includegraphics[width=\textwidth]{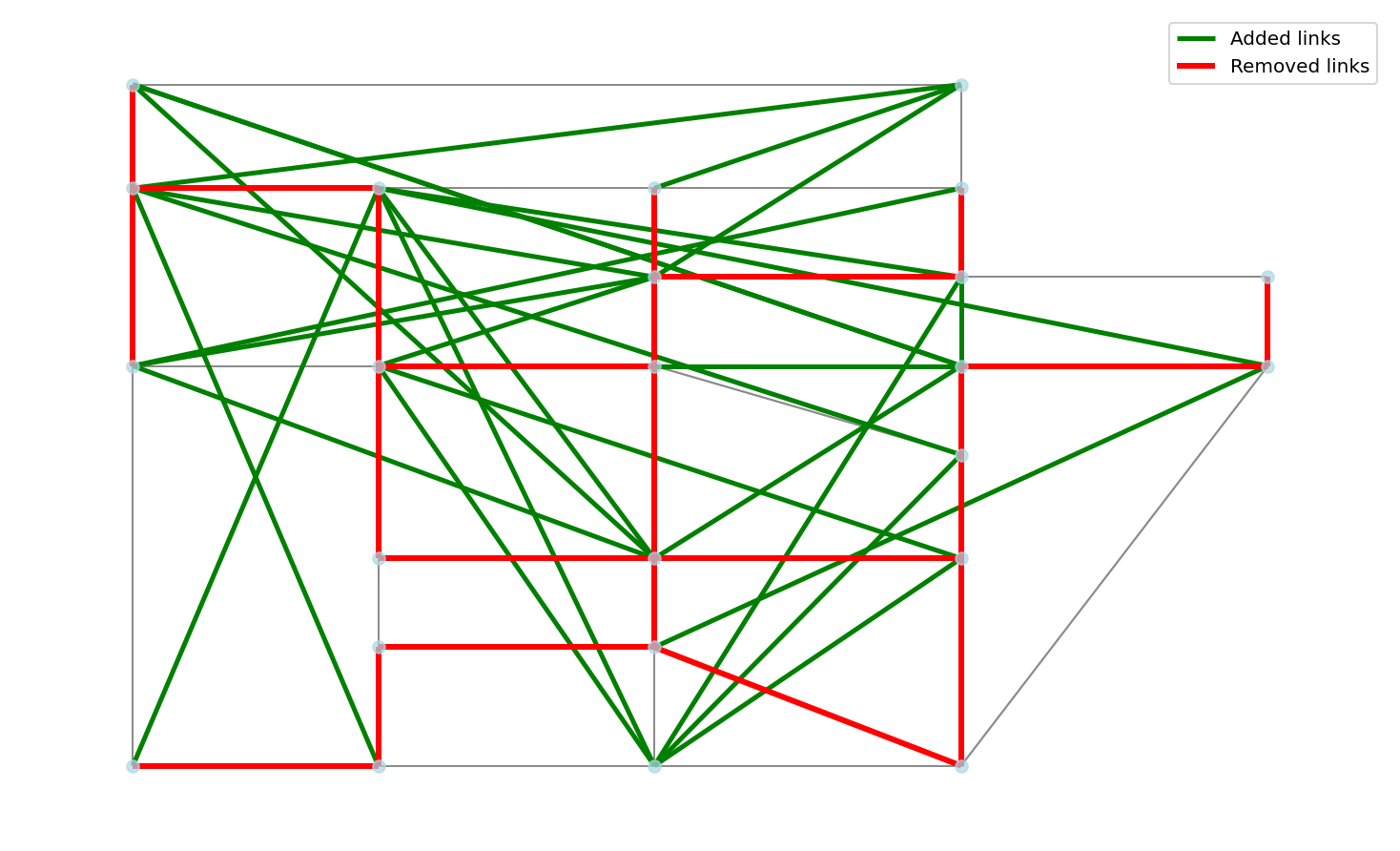}
        \caption{75\% variation}
        \label{fig:sf_topology_75}
    \end{subfigure}
    \caption{Illustrations of sampled generalized topological variations of the SiouxFalls network, showing increasing levels of structural change. Link colors indicate network modifications: red links correspond to removed links, while green links correspond to added links. }
    \label{fig:siouxfalls_variation}
\end{figure}

\begin{table}[H]
\centering
\caption{Model performance under different link-change fractions in topological variations}
\label{tab:link_change_performance}
\begin{tabular}{lcccc}
\toprule
\textbf{Link-change fraction} & \textbf{MAE} & \textbf{MAPE (\%)} & \textbf{Predicted Delay (\%)} & \textbf{UE Solution Delay (\%)} \\
\midrule
25\%  &17.11  &3.53  &0.60\%  &0.03\%  \\
50\%  &15.38  &3.13  &0.50\%  &0.03\%  \\
75\%  &31.06  &5.74  &0.64\%  &0.03\%  \\
\bottomrule
\end{tabular}
\end{table}

\noindent \textbf{Scenario 4:}

\newref{Table}{result-multi-class} presents the performance of the proposed model with a multi-class traffic network (cars and trucks). The MAPE for both vehicle types is low, particularly for trucks, indicating that the model's percentage error is relatively small, suggesting a generally good fit. The model predicts the path flow for trucks more accurately than for cars, as shown by the MAPE values: 1.55\% for trucks and 2.11\% for cars. The delay percentage based on the predicted path flow is also lower for trucks (0.35\%) compared to cars (0.36\%). In this scenario, link flow predictions are more accurate than path flow predictions (for cars, it is 1.02\% and 2.11\%, respectively).


\begin{table}[!ht]
\centering
\caption{Model performance with multi-class network}
\label{result-multi-class}
\begin{tabular}{lcc}
\hline
\textbf{Indicator} & \textbf{Car} & \textbf{Truck} \\ \hline
MAE & 4.10 & 1.95 \\
MAPE (\%) & 2.11 & 1.55 \\ 
Prediction delay (\%) & 0.36 & 0.35 \\ 
UE Solution delay (\%) & 0.17 & 0.09 \\  \hline
\end{tabular}
\end{table}

\vspace{2mm}

\noindent \textbf{Scenario 5:}

{\newref{Table}{tab:ema_anaheim_result} presents the performance of the proposed model when applied to the large-scale EMA and Anaheim network for predicting optimal path flow. The results demonstrate that even with a network of this scale, the model achieves a low error rate, with MAE of 6.01 and 1.43 respectiveley. The delay percentage of the network, based on the predicted path flow, is 2.36\% and 0.36\%, only 0.21\% and 0.36\% higher than the "ground-truth" solution. This indicates that the model's predictions are closely aligned with optimal UE conditions in terms of travel time delay. Furthermore, the proposed model achieves a prediction time of 0.003 seconds per OD matrix for the EMA network, representing approximately a 5,000-fold reduction compared to the 15-second solution time required by Gurobi. For the Anaheim network, the model attains a prediction time of 0.0017 seconds per OD matrix, which is more than 20,000 times faster than the 36-second solving time obtained using Gurobi.}


\begin{table}[H]
\centering
\caption{Model performance comparison between EMA and Anaheim networks (50\% OD missing)}
\label{tab:ema_anaheim_result}
\begin{tabular}{lcc}
\toprule
\textbf{Indicator} & \textbf{EMA Network} & \textbf{Anaheim Network} \\
\midrule
MAE & 6.01 & 1.43  \\
MAPE (\%) & 3.69 & 16.92 \\
Prediction delay percentage (\%) & 2.36 & 0.36 \\
UE Solution delay percentage (\%) & 2.15 &  0.0001\\
\bottomrule
\end{tabular}
\end{table}

For this complex problem, utilizing 4,000 samples provides an appropriate balance between avoiding overfitting and maintaining computational efficiency. A smaller sample size may increase the risk of overfitting, while a larger sample size could result in significantly higher computational costs. 

\section{Discussion and Conclusion}\label{sec:conclude}
This paper presents a novel framework for UE solution prediction in the traffic assignment problem utilizing the Transformer architecture, a state-of-the-art deep learning architecture known for its efficacy in handling complex sequence modeling tasks. The motivation for this research stems from the need to overcome the computational challenges associated with classical mathematical optimization methods for UE.
The proposed approach aims to predict path flows from the user's perspective, thus providing a more granular and theoretical solution. Importantly, the model learns path flows that inherently satisfy flow conservation and OD demand, attending to physical and operational constraints without requiring explicit enforcement during training. We do not consider all possible equilibrium path flows as reference points; therefore, even if our solution deviates from other equilibrium configurations, it may still achieve better overall quality according to the selected criteria.

To validate the proposed method, we conducted experiments on both synthetic networks and real-world urban transportation networks, considering various link-missing cases and multi-class scenarios with incomplete OD demand, and on a large-scale network, demonstrating the model's effectiveness in predicting optimal path flow distributions. The results demonstrate that the model can reliably predict path flows even when both OD demand data and network link data are missing. Moreover, it maintains robust performance under network topology changes without retraining, despite being trained only on original urban network with baseline missing OD demand. Thus, the framework not only reduces computational costs but also adapts to changes in network demand and structure without retraining, enabling robust `what-if' analyses for transportation planning and management. This study demonstrates the effectiveness of an attention-based sequence learning framework in capturing complex correlations between OD pairs (\newref{Figure}{fig:model-flow}), providing a robust surrogate model capable of accelerating complex optimization tasks in areas such as resource allocation and infrastructure management.



{Given the methodological focus of this study, and the differences in output representation and flow conservation treatment across existing models, a direct benchmark against state-of-the-art approaches was not pursued. At equilibrium, link flows are unique, but the corresponding path flows are not. Extending link-based models to predict path flows by modifying their final layer would not yield a uniquely defined or comparable reference, as multiple feasible path-flow sets can correspond to the same UE link-flow distribution. Such an extension would also require solving a linear program over all possible paths, leading to high computational complexity as network size grows.} Moreover, most existing studies do not provide publicly available code or reproducible implementations, making it challenging to ensure a fair and consistent comparison within the same path-flow framework. A systematic comparative evaluation in a unified path-flow setting remains an important direction for future work. 

{We note that as the number of OD pairs increases, both memory and computational requirements grow substantially due to the model operating over all OD-path combinations. In such large-scale settings, a practical solution is zone-based aggregation, where individual OD pairs are grouped into higher-level zones. This reduces the effective number of OD interactions and provides a coarser but computationally tractable representation of the network. There is therefore an inherent trade-off
between spatial resolution and computational scalability. Recent work 
has begun to address these scalability challenges directly. 
\citet{scalable_model_1, scalable_model_2} decompose large-scale network problems into parallel subproblems solved efficiently via GPU-accelerated solvers, demonstrating that significant computational gains are achievable without sacrificing solution quality. Integrating such approaches with the proposed framework represents a promising direction for future work.}

{Currently, the model learns and predicts static traffic flow patterns over a predefined path set and is not designed to handle scenarios where the predefined path set becomes infeasible due to structural network changes, such as link closures or capacity disruptions. While training on augmented disruption scenarios is a conceptually straightforward extension, it introduces a significant data generation burden, as each disruption scenario requires regenerating feasible path 
sets and re-solving the traffic assignment problem to obtain ground truth flows. This makes large-scale augmentation computationally expensive. Consequently, a path-agnostic encoding based on intrinsic path features such as travel time, number of links or path length among others represents a more practical and scalable architectural direction, as it may improve robustness to path set changes without requiring additional ground truth data.}
As another potential extension, the framework could be adapted to predict dynamic traffic flow patterns by incorporating time-dependent path flows.

\section*{Acknowledgments}

This work was supported by the France–Berkeley Fund under the project "MADyNE" and the French ANR research project SMART-ROUTE under Grant ANR-24-CE22-7264.


\bibliography{biblio1}

\appendix
\section{Sensitivity check for Average Delay normalization} \label{AppA}

The numerator in the Average Delay (AD) formulation given by Equation \ref{Equation16}, gives the excess travel time relative to the equilibrium condition and the normalization by total demand converts AD into a per-unit-demand delay. Small AD values could arise either because the excess travel time approaches zero or because of a large denominator in the AD formula. To ensure that the normalization does not unintentionally downplay deviations from equilibrium, we report the distribution of magnitude of the excess cost relative to mean travel times, per-OD pair. 

For each OD pair $r$ and mode $z$, the percentage excess cost on path $p$ is defined as:

\begin{equation}
\mathrm{ExcessCost}^{\,r,z}_{p} = \frac{c^{\,r,z}_{p} - u^{\,z}_{r}}{\bar{c}^{\,z}} \times 100,
\end{equation}

where $c^{\,r,z}_{p}$ is the path cost, $u^{\,z}_{r} = \min_{p} c^{\,r,z}_{p}$ is the minimum path cost for that OD pair, and $\bar{c}^{\,z}$ is the network-wide mean path cost.

The \newref{Figure}{fig:errors_sf} and \newref{Figure}{fig:errors_ah} present the histogram of delay distribution on path 1 for single class, across all OD pairs of all test matrices in the SiouxFalls and Anaheim network respectively. The histogram reveals that close to 80\% of OD pairs in SiouxFalls and more than 80\% for Anaheim have percentage excess cost values close to zero, indicating that most flows follow the shortest path with costs nearly identical to their respective minimum path costs. This figure confirms that the low average delay values are indeed due to the numerators being close to zero, in contrast to being a result of large total demand values in the denominator.

 \setcounter{figure}{0}
\begin{figure}[H]
    \centering
    \begin{subfigure}[b]{0.48\textwidth}
        \centering
        \includegraphics[width=\textwidth]{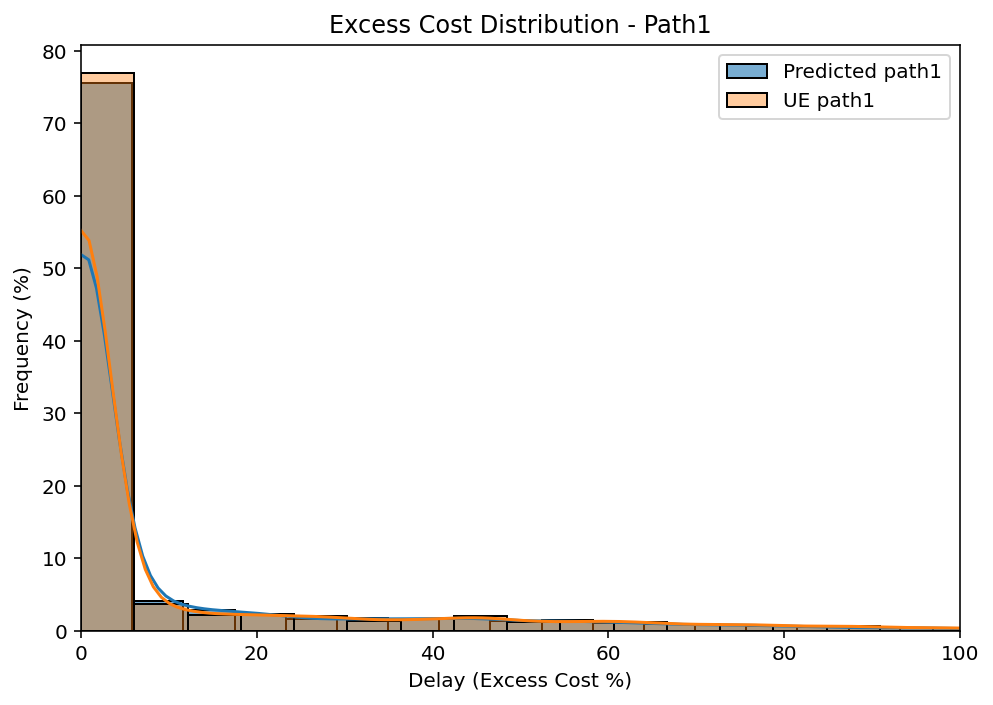}
        \caption{SiouxFalls network}
        \label{fig:errors_sf}
    \end{subfigure}
    \hfill
    \begin{subfigure}[b]{0.48\textwidth}
        \centering
        \includegraphics[width=\textwidth]{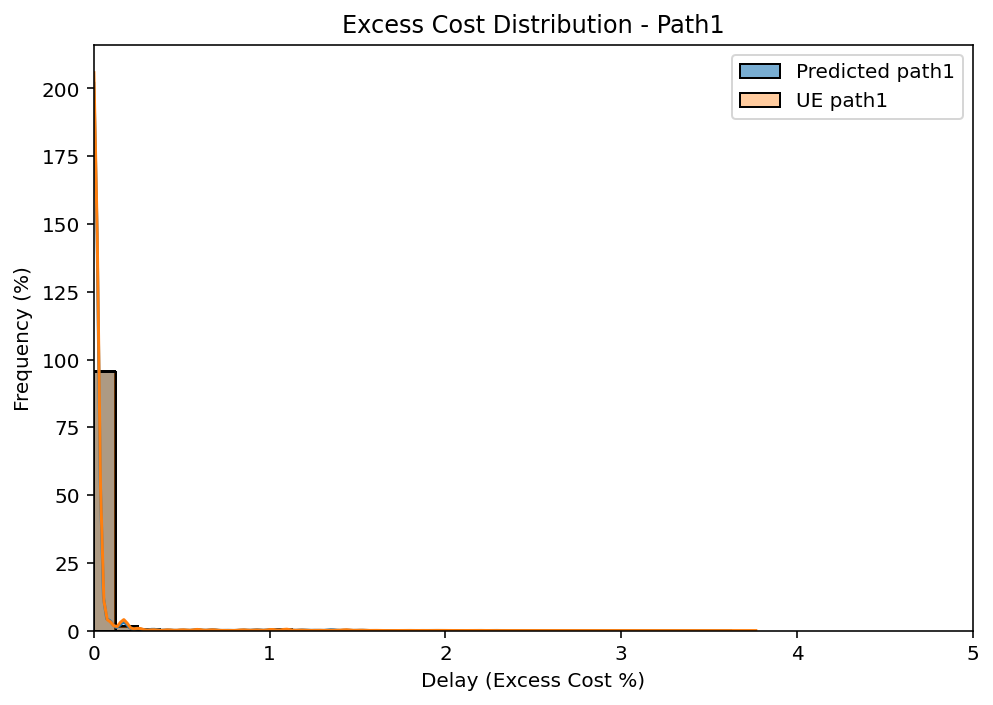}
        \caption{Anaheim network}
        \label{fig:errors_ah}
    \end{subfigure}
    \caption{Histograms of Excess cost percentage across all OD pairs \(r\) for class \(z\) for (a) SiouxFalls and (b) Anaheim networks.}
\end{figure}

\section{Check for Relative OD Conservation Error} \label{error_od}


{The empirical results presented in Figure report the relative OD conservation error. The relative OD conservation error measures the normalized deviation between the predicted total flow assigned to each OD pair and its corresponding demand, aggregated over all OD pairs. We observe that the deviations remain less than 2\% for all test OD matrix samples.}

\begin{equation}
\epsilon_{\text{OD}} =
\frac{\sum_{r \in R} \left| \sum_{p=1}^{k} f_{r,k} - x_r \right|}
{\sum_{r \in R} x_r},
\end{equation}
where $R$ denotes the set of origin--destination pairs, $k$ the number of paths per OD pair, $f_{r,p}$ the predicted flow on path $p$ for OD pair $r$, and $x_r$ the corresponding OD demand.

\setcounter{figure}{0}
\begin{figure}[H]
    \centering
    \includegraphics[width=1\textwidth]{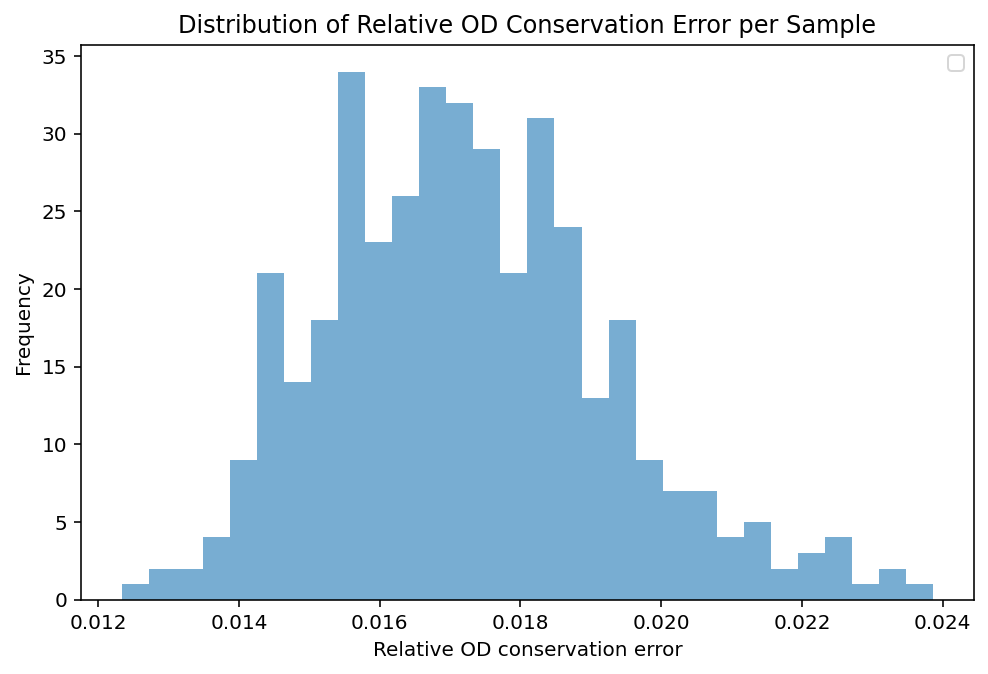}
    \caption{Relative OD Conservation Error per Test Matrix}
    \label{fig:errors3}
\end{figure}

\end{document}